\documentclass{article}

\usepackage{microtype}
\usepackage{graphicx}
\usepackage{subfigure}
\usepackage{amsmath}
\usepackage{multirow}
\usepackage{amssymb}
\usepackage{bbding}
\usepackage{float}
\usepackage{caption}
\usepackage{stfloats}

\usepackage{mwe}
\usepackage{booktabs}
\usepackage{threeparttable}
\usepackage{bm}
\usepackage{colortbl}
\usepackage{hyperref}

\newcommand{\ib}[2]{$\text{#1}_{\uparrow \text{#2}\%}$} 
\newcommand{\db}[2]{$\text{#1}_{\downarrow \text{#2}\%}$} 

\usepackage{colortbl}
\definecolor{grayDark}{gray}{0.9}
\definecolor{grayLight}{gray}{0.98}

\DeclareMathOperator*{\argmin}{arg\,min}
\usepackage{booktabs} 

\usepackage{hyperref}

\usepackage[accepted]{icml2022}

\icmltitlerunning{SDQ: Stochastic Differentiable Quantization with Mixed Precision}

\begin{document}

\twocolumn[
\icmltitle{SDQ: Stochastic Differentiable Quantization with Mixed Precision}




\begin{icmlauthorlist}
\icmlauthor{Xijie Huang}{ust}
\icmlauthor{Zhiqiang Shen}{ust,mbzuai,IAS}
\icmlauthor{Shichao Li}{ust}
\icmlauthor{Zechun Liu}{meta}
\icmlauthor{Xianghong Hu}{ust,access}
\icmlauthor{Jeffry Wicaksana}{ust}
\\
\icmlauthor{Eric Xing}{cmu,mbzuai}
\icmlauthor{Kwang-Ting Cheng}{ust}
\end{icmlauthorlist}

\icmlaffiliation{ust}{Hong Kong University of Science and Technology}
\icmlaffiliation{cmu}{Carnegie Mellon University}
\icmlaffiliation{mbzuai}{Mohamed bin Zayed University of Artificial Intelligence}
\icmlaffiliation{meta}{Reality Labs, Meta Inc}
\icmlaffiliation{IAS}{Jockey Club Institute for Advanced Study, HKUST}
\icmlaffiliation{access}{ACCESS - AI Chip Center for Emerging Smart Systems}
\icmlcorrespondingauthor{Zhiqiang Shen}{zhiqiangshen0214@gmail.com}

\icmlkeywords{Machine Learning, ICML}

\vskip 0.3in
]



\printAffiliationsAndNotice{}  

\begin{abstract}
In order to deploy deep models in a computationally efficient manner, model quantization approaches have been frequently used. 
In addition, as new hardware that supports mixed bitwidth arithmetic operations, recent research on mixed precision quantization (MPQ) begins to fully leverage the capacity of representation by searching optimized bitwidths for different layers and modules in a network. 
However, previous studies mainly search the MPQ strategy in a costly scheme using reinforcement learning, neural architecture search, etc., or simply utilize partial prior knowledge for bitwidth assignment, which might be biased on locality of information and is sub-optimal.
In this work, we present a novel \textbf{S}tochastic \textbf{D}ifferentiable \textbf{Q}uantization \textbf{(SDQ)} method that can automatically learn the MPQ strategy in a more flexible and globally-optimized space with smoother gradient approximation. 
Particularly, Differentiable Bitwidth Parameters (DBPs) are employed as the probability factors in stochastic quantization between adjacent bitwidth choices. 
After the optimal MPQ strategy is acquired, we further train our network with Entropy-aware Bin Regularization and knowledge distillation.
We extensively evaluate our method for several networks on different hardware (GPUs and FPGA) and datasets. 
SDQ outperforms all state-of-the-art mixed or single precision quantization with a lower bitwidth and is even better than the full-precision counterparts across various ResNet and MobileNet families, demonstrating its effectiveness and superiority.
\footnote{Project: \url{https://huangowen.github.io/SDQ/}.}
\end{abstract}

\section{Introduction}

\begin{figure}[t]
\centering
\subfigure[Proposed Stochastic Differentiable Quantization (SDQ)]{%
\label{Figure:intro}%
\includegraphics[width=0.5\textwidth]{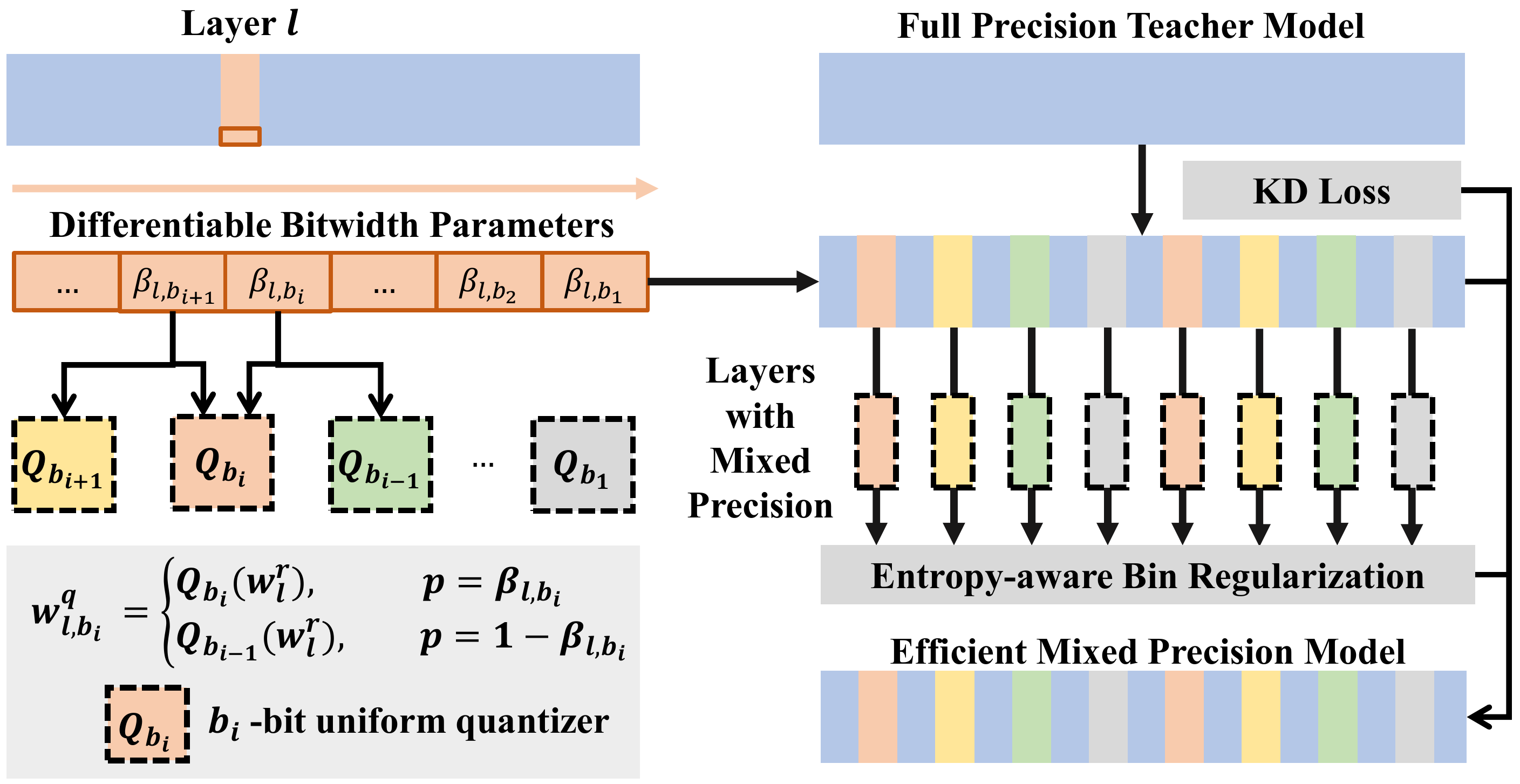}}
\subfigure[Full Precision]{%
\label{fig:3dloss-fp}%
\includegraphics[width=0.17\textwidth]{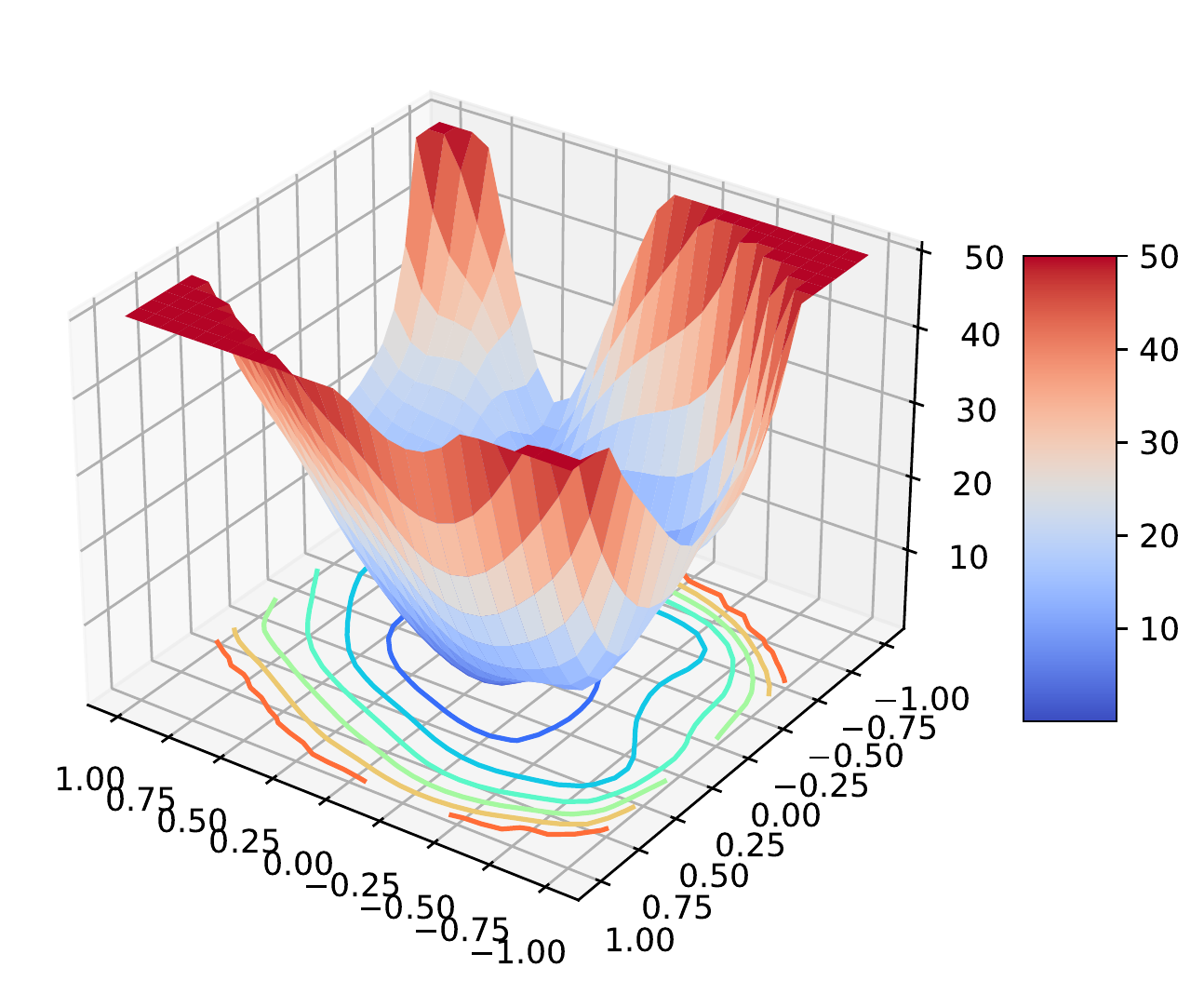}}%
\subfigure[Interpolation]{%
\label{fig:3dloss-linear}%
\includegraphics[width=0.15\textwidth]{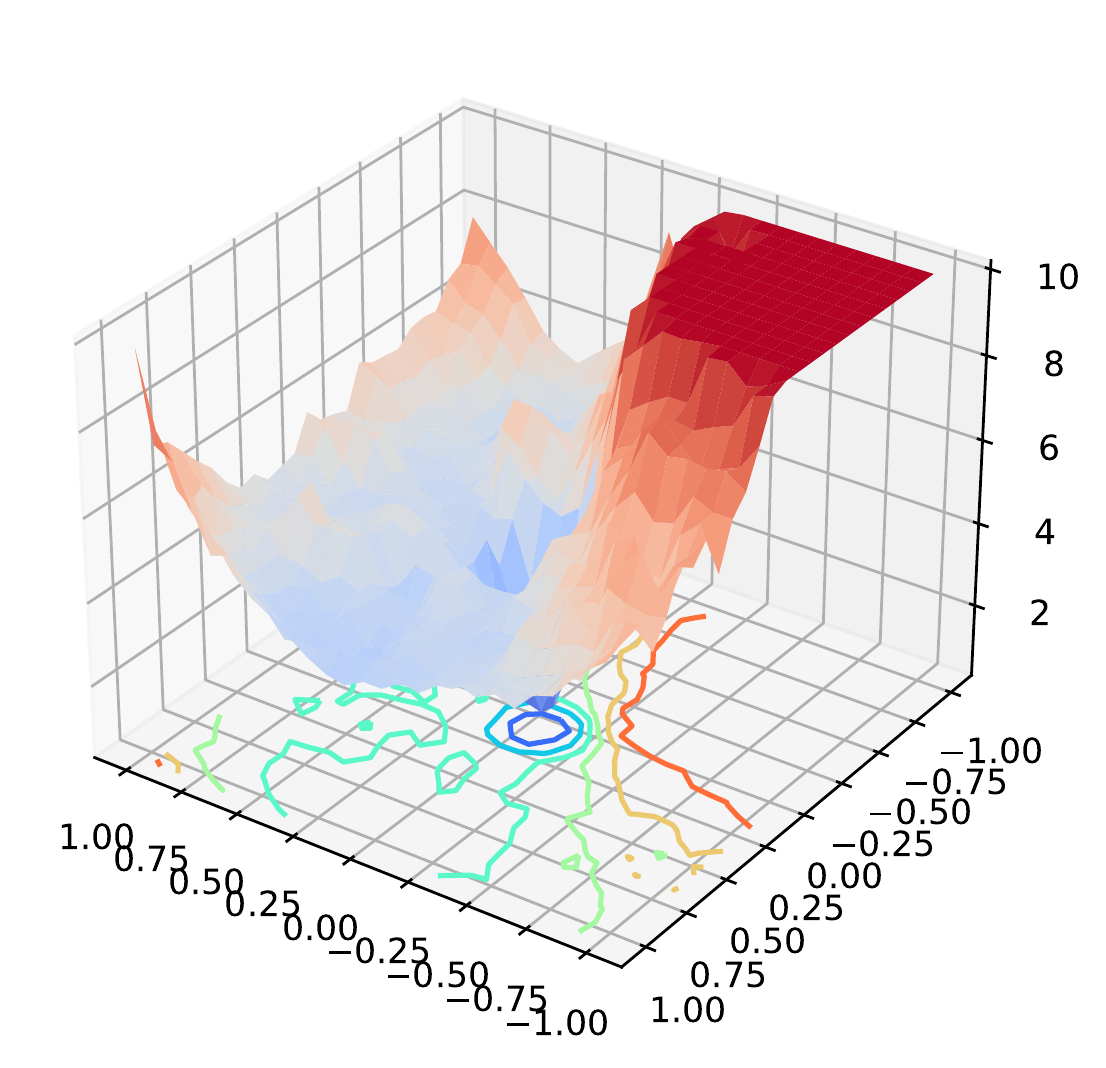}}%
\subfigure[Stochastic]{%
\label{fig:3dloss-ours}%
\includegraphics[width=0.17\textwidth]{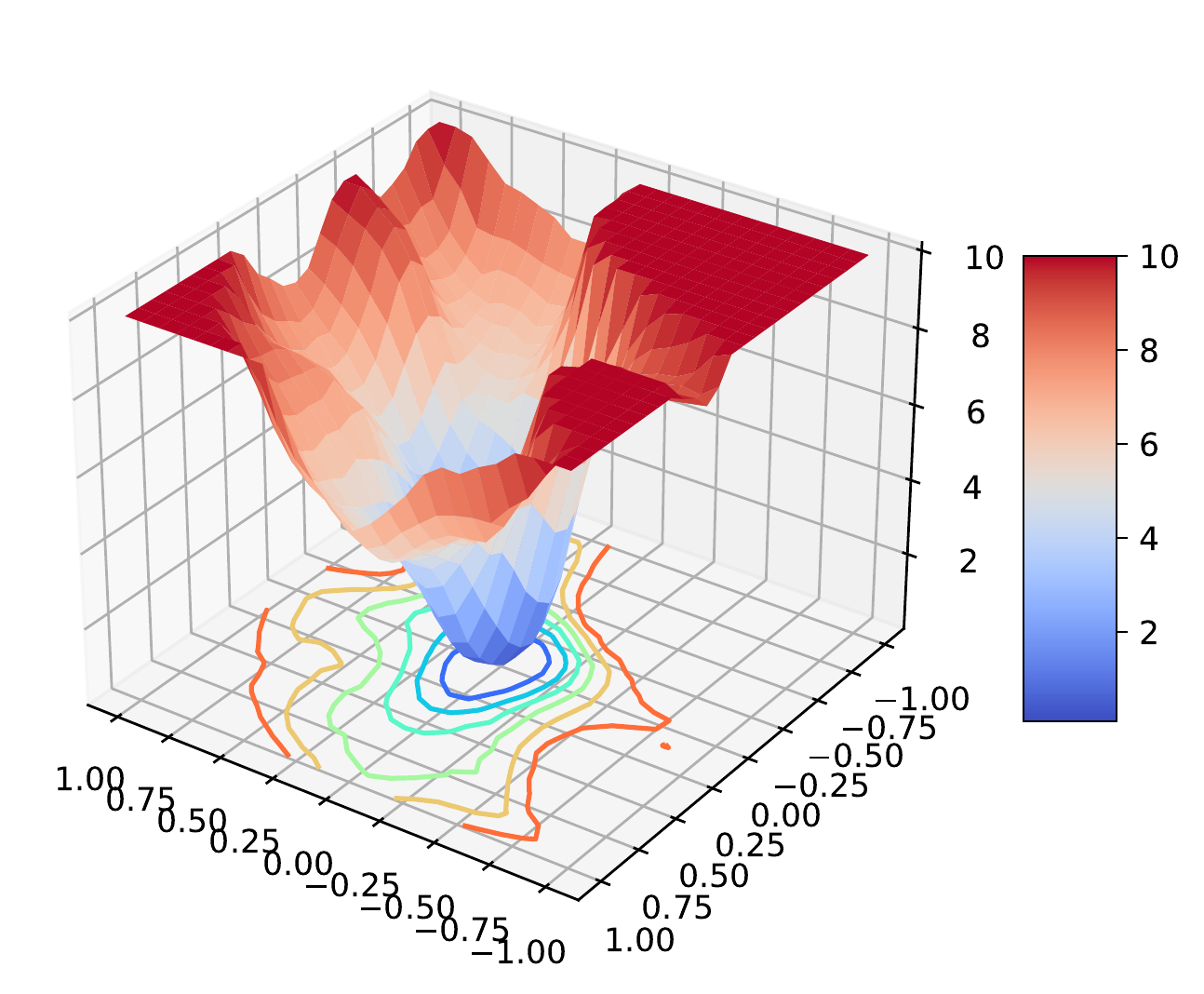}}%
\caption{\ref{Figure:intro} demonstrates how our SDQ generates optimal MPQ strategy and train a MPQ network effectively. \ref{fig:3dloss-fp},~\ref{fig:3dloss-linear},~\ref{fig:3dloss-ours} compare the underlying loss optimization landscape~\cite{li2018visualizing} of a full precision model, MPQ of ResNet20 with linear interpolation~\cite{yang2021fracbits, nikolic2020bitpruning}, and MPQ of ResNet20 with stochastic quantization.}
\end{figure}

Deep learning models such as Convolutional Neural Networks (CNNs) have demonstrated outstanding performance on various tasks, {\em e.g.,} visual recognition~\cite{krizhevsky2012imagenet, he2016deep, tan2019efficientnet}. Their latency, energy consumption, and model size have become the most significant factors for consideration of the deployment with these deep learning models. 
In recent years, researchers have studied the design, training, and inferencing techniques of deep learning models for greater computation efficiency, including compact network design and search~\cite{howard2017mobilenets, pham2018efficient, guo2020single}, knowledge distillation~\cite{hinton2015distilling}, pruning~\cite{liu2017learning, liu2018rethinking}, quantization~\cite{zhou2016dorefa, zhang2018lqnet}, and sparsity~\cite{wen2016learning}.
Quantization techniques have attracted particular attention because emerging hardware accelerators~\cite{judd2016stripes, jouppi2017datacenter, sharma2018bit} begin to support low-precision inference.

To fully exploit the difference of representative capacity and redundancy in different parts (blocks, layers, and kernels) of deep neural networks, mixed precision quantization (MPQ) techniques have been put forward to assign different bitwidths to various components in a network. 
To achieve an MPQ network with satisfactory accuracy and compression, two critical problems must be solved consecutively: First, given a pre-trained network with weights $\{\bm W ^{(l)}\}_{l=1}^{L}$ and a training dataset $\mathcal{\bm D}$, how to find an optimal quantization strategy $\{\bm b^{(l)} \}_{l=1}^{L}$? Prior solutions addressed this general goal for MPQ are primarily based on searching or optimization. 

Whereas, the search-based methods which conduct an iterative scheme to explore the best structure cannot globally optimize the MPQ network for all modules and are also expensive to learn\footnote{For $L$ layers with $B$ bitwidth candidate, searching both layer-wise weights and activations has a time complexity of $\mathcal{O}(B^{2L})$.}. Meanwhile, existing optimization-based methods, such as FracBits~\cite{yang2021fracbits}, suffer from unstable optimization as the approximation design for gradients is not smooth and stable.  
Consequently, an ideal solution should be differentiable which enables automatic optimization of the bitwidth assignment in a global optimization space with more precise gradient approximations and smoother latent landscape as illustrated in Figure~\ref{fig:3dloss-ours}, which derives the primary motivation of this work.

The second problem stems from the quantization-aware post-training: after the quantization strategy $\{\bm b^{(l)} \}_{l=1}^{L}$ is acquired, how can we train a quantized network $\{\bm W ^{(l)} _{\bm b^{(l)}}\}_{l=1}^{L}$ for maximizing the potential of performance and minimizing the accuracy sacrifice compared to a full-precision model? Empirically, during quantization training, the input information should be preserved and quantization error should be minimized through proper loss function design.

To address the above problems, in this work, we propose a novel \textbf{S}tochastic \textbf{D}ifferentiable \textbf{Q}uantization \textbf{(SDQ)} to tackle the challenges of searching and training MPQ. 
Concretely, we present a one-shot solution via representing the choice of discrete bitwidths as a set of Differentiable Bitwidth Parameters (DBPs), as shown in Fig.~\ref{Figure:intro}. DBPs are utilized in the forward path of the network as probability factors during the \textbf{stochastic quantization}. During the quantization strategy generation, DBPs will be updated to learn the optimal bitwidth assignment automatically. During the backpropagation, we use Gumbel-softmax reparameterization to estimate the gradient so the gradient can be back-propagated smoothly through our DBPs. Compared to previous differentiable solutions~\cite{nikolic2020bitpruning, yang2021fracbits} using linear interpolation, the stochastic quantization scheme can substantially improve the network training stability and help the DBPs converge with a smoother loss landscape shown in Fig.~\ref{fig:3dloss-ours}. 

Addressing the challenge imposed by the second problem, we propose an Entropy-aware Bin Regularizer (EBR) based on entropy analysis, which regularizes the weights to keep more information-carrying components while considering the various precision of different layers. Knowledge distillation is also used to fully leverage the representation capability of full-precision teacher models.

We demonstrate the advantage of our SDQ in terms of effectiveness and efficiency on different networks and datasets. For ResNet18 on the ImageNet-1K dataset, our quantized model can significantly improve top-1 accuracy (71.7\%) compared to the full precision model (70.5\%) with average bitwidth of both weights and activations no more than 4 bits. We also deploy ResNet18 on Bit Fusion~\cite{sharma2018bit} accelerator to show the efficiency of our SDQ models.\\
In summary, our contribution can be concluded as:
\vspace{-0.1in}
\begin{itemize}
    \addtolength{\itemsep}{-0.08in}
    \item We present a novel SDQ framework to learn the optimal mixed precision quantization strategy via a set of differentiable bitwidth parameters as probability factors during the stochastic quantization. 
    \item We utilize straight-through Gumbel-softmax estimator in the gradient computation \textit{w.r.t.} differentiable bitwidth parameters. We also incorporate the quantization error regularization term while learning the mixed precision quantization strategy. 
    \item We propose an Entropy-aware Bin Regularizer to minimize the quantization error while considering mixed precision, which helps preserve more information-carrying components.
    \item We extensively evaluate our method on different network architectures and datasets. We further conduct {\bf deployment experiments} on various hardware (GPUs and a real FPGA system) to demonstrate the effectiveness and efficiency of our models. 
\end{itemize}

\section{Related Work}

Quantization techniques can be categorized into uniform and non-uniform quantization based on the quantization intervals. Non-uniform quantization~\cite{miyashita2016convolutional, zhang2018lqnet, li2019apot}, due to its flexible representation, usually can better allocate the quantization values to minimize the quantization error and achieve better performance than uniform schemes. 
In addition, the quantization methods can also be classified as stochastic and deterministic quantization. Inspired by gradient estimation of stochastic neurons~\cite{bengio2013estimating}, stochastic quantization (SQ)~\cite{courbariaux2015binaryconnect, courbariaux2016binarized, dong2019stochastic} has been explored. Unlike previous deterministic quantization, real-value variables of SQ are quantized to different quantization levels controlled by a probability distribution. For example, BinaryConnect~\cite{courbariaux2015binaryconnect} transforms the real-value weights into +1 or -1 with probability determined by the distance to the zero point.

However, the aforementioned quantization schemes solely assign the same bitwidth to the entire model. Mixed precision quantization (MPQ), which employs different bitwidths in distinct layers or modules, can achieve a higher compression ratio and improved model capabilities. MPQ is a more promising direction in general, and it usually is divided into three categories: \textbf{Search-Based Methods}, \textbf{Metric-Based Methods}, and \textbf{Optimization-Based Methods}. 

\textbf{Search-Based Methods} usually utilize Neural Architecture Search~\cite{wu2018mixed, yu2020bpnas, guo2020single} or Reinforcement Learning~\cite{wang2019haq, elthakeb2019releq} to perform  searching of quantization strategies. For instance, HAQ~\cite{wang2019haq} leverages reinforcement learning and incorporates hardware feedback in the loop. DNAS~\cite{wu2018mixed} explore the quantized search space with gradient-based optimization to improve the efficiency. Despite these efforts to increase the efficiency of searching, the time and computational cost of generalizing search-based approaches to various network designs, datasets, and hardware platforms remain impediments.

\textbf{Metric-Based Methods} target at assigning bitwidth considering specific metrics that can be computed easily. HAWQ~\cite{dong2019hawq, dong2019hawqv2, yao2021hawqv3} generates MPQ strategy based on the layers' Hessian spectrum. OMPQ~\cite{ma2021ompq} utilizes layer orthogonality to construct a linear programming problem to derive the bitwidth configuration. SAQ~\cite{liu2021sharpness} determines the bitwidth configurations of each layer, encouraging lower bits for layers with lower sharpness. Although these methods are highly computation efficient, the generated MPQ strategies are usually sub-optimal as the mapping from these metrics to the bitwidths is manually established.

\textbf{Optimization-Based Methods} formulate the MPQ strategy generation problem as an optimization problem. The core challenge is to tackle the non-differentiability of task loss \textit{w.r.t.} the bitwidth assignment. FracBits~\cite{yang2021fracbits} proposed a fractional bitwidth parameter and used linear interpolation during the forward of quantization. DDQ~\cite{zhaoyang2021DDQ} proposed a method to decompose the quantizer operation into the matrix-vector product. Uhlich et al.~\cite{uhlich} proposed differentiable quantization via parametrization. These methods either introduce extra parameters such as quantization levels and the dynamic ranges of quantizers as the optimization target, or utilize linear interpolation between different quantization levels. The interpolation operation will introduce a large number of useless optimization targets for quantization, and more instability is expected. 

Noticeably, our work falls into the categories of \textbf{uniform}, \textbf{quantization-aware training}, and \textbf{optimization-based methods} of mixed precision network. To the best of our knowledge, SDQ is the first quantization strategy that adopts stochastic quantization to optimize the bitwidth assignment. Compared to previous research which also leverages the idea of differentiable bitwidth such as FracBits~\cite{yang2021fracbits} and BitPruning~\cite{nikolic2020bitpruning}, SDQ provides better training stability compared to the naive linear interpolation. Our DBPs can further denote discrete bitwidth candidates to better match the configuration of particular hardware accelerators, while others cannot, {\em e.g.,} Bit Fusion~\cite{sharma2018bit} only supports powers of 2 bitwidth. 

\section{Method}
\subsection{Preliminaries}
In network quantization, real-valued weights $\bm w^r$ and activations $\bm x^r$ are converted to low-precision value $\bm w^q$ and $\bm x^q$. To mitigate the problem that gradient-based optimization cannot be directly applied to quantized value, previous research like DoReFa-Net~\cite{zhou2016dorefa} exploits a straight-through estimator (STE)~\cite{bengio2013estimating} to assign the gradients passing in $\bm x^{\bf i}$ and out $\bm x^{\bf o}$ to be equal. Assume the loss function for the model is $\mathcal{\bm L}$, an STE for $b$-bit uniform quantizer $\bm q_b$ can be denoted as:
\begin{equation}
\begin{aligned}
\label{equ:dorefa}
\resizebox{0.42\textwidth}{!}{$\begin{split}
\text{\textbf{Forward:}}&~~\bm x^{\bf o} = \bm q_b(\bm x^{\bf i})=\frac{1}{2^b-1} \text{round}[(2^b-1)\bm x^{\bf i}];~~ \\ 
\text{\textbf{Backward:}}&~~\frac{\partial \mathcal{\bm L}}{\partial \bm x^{\bf o}}=\frac{\partial \mathcal{\bm L}}{\partial \bm x^{\bf i}}.
\end{split}$}
\end{aligned}
\end{equation}

Here $\bm x^{\bf i}, \bm x^{\bf o} \in [0,1]$. The weights and activations are first linearly transformed and clamped to interval $[0,1]$. The complete scheme for $b$-bit uniform quantizer $\bm Q_b$ is:
\begin{equation}
\label{equ:dorefa-quant}
\resizebox{0.4\textwidth}{!}{$\begin{split}
    \bm x^q = \bm Q_{b}(\bm x^r) = \bm q_b(\frac{\tanh(\bm x^r)}{2 \max(|\tanh(\bm x^r)|)} + \frac12) - 1 \text{.}
\end{split}$}
\end{equation}

\subsection{Generating Quantization Strategy}
In the quantization strategy generation phase, our goal is to learn the optimal bitwidth assignment. We introduce Differentiable Bitwidth Parameters (DBPs) $\{\bm \beta_{l, b_i}\}$ for each layer $l \in L$ and bitwidth candidate $b_i \in \mathcal B$, where $i$ is the index of bitwidth in the candidate set $\mathcal B$. DBPs are initialized to 1 to represent a deterministic quantization at different levels. \\
During forward, the weights can be quantized to two different bitwidths: the current bitwidth $b_i$ and next bitwidth $b_{i-1}$ in the candidate sets. The probability of quantization into two bitwidths is controlled by the DBP $\bm \beta_{l, b_i}$ at this layer.
\begin{equation}
\label{eq:prob_quant}
\resizebox{0.45\textwidth}{!}{$\begin{split}
{\bm w}^{q}_{l,b_i} = \left\{
             \begin{array}{ll}
             \vspace{0.5em} 
             \!\bm Q_{b_i}(\bm w_l^r) & \text{with probability} \ \ \bm p = \bm \beta_{l, b_i} \\
             \!\bm Q_{b_{i-1}}(\bm w_l^r) & \text{with probability} \ \ \bm p = 1-\bm \beta_{l, b_i}
             \end{array}
\right.,
\end{split}$}
\end{equation}
where ${\bm w}^{q}_{l,b_i}$ represents the quantized weight under $b_i$-bit stochastic-bitwidth quantizer. According to the characteristic of STE introduced in Eq.~\ref{equ:dorefa}, the gradient through the quantization values and real values are the same: $\frac{\partial \mathcal{L}}{\partial {\bm w}^{r}_{l}}=\frac{\partial \mathcal{L}}{\partial Q_{b_i}(\bm w_l^r)} = \frac{\partial \mathcal{L}}{\partial Q_{b_{i-1}}(\bm w_l^r)}.$ Thus, the expected gradients of quantized weight ${\bm w}^{q}_{l,b_i}$ over all trainable parameters in the network can be computed as:
\begin{equation}
\label{eq:expect_grad_w}
\begin{aligned}
\resizebox{0.5\textwidth}{!}{$\begin{split}
\mathbb{E}[\frac{\partial  \mathcal{L}}{\partial {\bm w}^{q}_{l,b_i}}] &= 
\bm \beta_{l, b_i}\mathbb{E}[\frac{\partial \mathcal{L}}{\partial Q_{b_i}(\bm w_l^r)} ] + 
(1-\bm \beta_{l, b_i})\mathbb{E}[\frac{\partial \mathcal{L}}{\partial Q_{b_{i-1}}(\bm w_l^r)}]  = \mathbb{E}[\frac{\partial \mathcal{L}}{\partial {\bm w}^{r}_{l,b_i}}],
\end{split}$}
\end{aligned}
\end{equation}
which is the same as the conventional STE. The derived gradient shows the proposed DBPs will not influence the gradient-based optimization of weight parameters during the training. 

However, the real gradient still cannot backpropagate directly through the probability parameter $\bm \beta_{l, b_i}$. To tackle this problem, we use {\em Straight-Through Gumbel SoftMax Estimator}~\cite{jang2016categorical} as a reparameterization trick to allow gradients flow from quantized value ${\bm w}^{q}_{l,b_i}$ to DBP $\bm \beta_{l, b_i}$. We can define the forwarding of previous stochastic-bitwidth quantization as ${\bm w}^{q}_{l,b_i}=\bm c_{l,b_i}^k \bm Q_{b_i}(\bm w_l^r) + (1-\bm c_{l,b_i}^k) \bm Q_{b_{i-1}}(\bm w_l^r)$, where $\bm c_{l,b_i}^k\!\! \in\! \{0,1\}$ is a choice variable and it follows a Bernoulli distribution $\bm c_{l,b_i}^k \sim \text{\em Bernoulli}(\bm \beta_{l, b_i})$. We use Gumbel SoftMax to transform the choice variable into a ``soft'' sampling operation while maintaining the distribution characteristics via:
\begin{equation}
\resizebox{0.48\textwidth}{!}{$\begin{split}
\bm c_{l,b_i}^k = \frac{\text{exp}((\log(\bm \beta_{l, b_i})+\bm g^k)/\bm \tau)}{ \text{exp}((\log(\bm \beta_{l, b_i})+\bm g^{k_0})/\bm \tau) + \text{exp}((\log(1-\bm \beta_{l, b_i})+\bm g^{k_1})/\bm \tau)}, 
\end{split}$}
\end{equation}
where $\bm g^k$, $\bm g^{k_0}$ and $\bm g^{k_1}$ are samples drawn from Gumbel(0,1) distribution, and $\bm \tau$ is the temperature coefficient to control the generated distribution. As has been proven in previous research~\cite{jang2016categorical}, the gradient now can be easily computed and the expected value of gradient \textit{w.r.t.} quantized weights given in Eq.~\ref{eq:expect_grad_w} still holds.

The advantage of MPQ is that different sensitivity of each layer to the quantization perturbation is fully considered. Previous research~\cite{dong2019hawq} has shown that this sensitivity is closely aligned to the number of parameters and quantization error of this layer. In light of this, we propose a quantization-error regularizer to penalize layers with a large number of parameters which result in high quantization error:
\vspace{-0.05in}
\begin{equation}
\label{eq:qer}
\mathcal{L}_{QER} = \sum_{l}^{L} \bm \beta_{l, b_i} \lambda_{b_i}||{\bm w}^{q}_{l,b_i}-\bm w_{l}^r||_2^2, 
\vspace{-0.05in}
\end{equation}
where $\lambda_{b_i}$ is a weighting coefficient to balance between different bitwidth, and $||{\bm w}^{q}_{l,b_i}-\bm w_{l}^r||_2^2$ is the $L_2$-norm of quantization error. This term is highly related to the bitwidth as it increases exponentially with the bitwidth. In our practice, we use $\lambda_{b_i} = (2^{|b_i|}-1)^2$ to balance between different precision (please see Appendix~\ref{sec:qer-analysis} for more details). We only optimize DBPs and will not optimize weight parameters with this quantization-error regularization. The complete optimization objective in this stage can be formulated as:
\begin{equation}
\begin{aligned}
    \mathcal{O} &= \argmin_{\bm W, \bm \beta} \mathcal{L}_{task} + \argmin_{\bm \beta} \lambda_{Q} \mathcal{L}_{QER},
\end{aligned}
\end{equation}
where $\mathcal{L}_{task}$ is the task loss and $\lambda_{Q}$ is the coefficient to balance between task loss and quantization error regularization.

During the quantization strategy generation phase, we start from the highest precision in the bitwidth candidate sets, and progressively reduce the bitwidth. When DBP at layer $l$ satisfies the condition $\bm \beta_{l, b_i} < \bm \beta_{\text{t}}$, where $\bm \beta_{\text{t}}$ represents the pre-defined DBP threshold, the bitwidth assignment for layer $l$ will be reduced from $b_i$ to $b_{i-1}$ and we will start to optimize $\bm \beta_{l, b_{i-1}}$. After a pre-defined number of epochs for training, we can generate an MPQ strategy $b_{l}$ for each layer $l \in L$ based on the current DBPs $\{\bm \beta_{l, b_0}, \cdots \bm \beta_{l, b_n}\}$ that have been optimized.

\subsection{Training with Quantization Strategy}
While prior MPQ research~\cite{uhlich, zhaoyang2021DDQ} combined the searching and training stages, we choose a two-stage strategy since the optimization objectives are distinct in these two periods.
In addition, more instability of training is expected when more quantization parameters are optimized simultaneously (\textit{i.e.} DBPs) with the weight parameters. In light of this, we apply quantization-aware training with the generated MPQ strategy and without the DBPs. The total loss $\mathcal{L}$ for optimization in this stage is:
\begin{equation}
\begin{aligned}
\resizebox{0.2\textwidth}{!}{$\begin{split}
    \mathcal{L} = \mathcal{L}_{KD} + \lambda_{E} \mathcal{L}_{EBR},
\end{split}$}
\end{aligned}
\end{equation}
where $\mathcal{L}_{KD}$ denotes the knowledge distillation loss and $\mathcal{L}_{EBR}$ denotes the entropy-preserving bin regularization. $\lambda_{E}$ is the weighting coefficient to balance between them. We will introduce them in Sec.~\ref{sec:kd} and Sec.~\ref{sec:ebr}, respectively.

\subsubsection{Knowledge Distillation}\label{sec:kd}
Intrinsically, a quantized classification network should learn an ideal similar mapping from input images to the output logits as a full-precision network, and the performance gap between them needs to be minimized. Based on this insight, we use knowledge distillation to train our MPQ network with a full-precision model as the teacher. The loss function is designed to enforce the similarity between the output distribution of full-precision teacher and MPQ student model:
\begin{equation}
\label{eq:kd}
\begin{aligned}
\resizebox{0.35\textwidth}{!}{$\begin{split}
    \mathcal{L}_{KD} = -\frac{1}{N}\sum_c\sum^N_{i=1} p_c^{\mathcal{F}_{\theta}}(X_i)\log(p_c^{\mathcal{Q}_{\theta}}(X_i))
\end{split}$}
\end{aligned}
\end{equation}
where the KD loss is defined as the cross-entropy between the output distributions $p_c$ of a full-precision teacher $\mathcal{F}_{\theta}$ and a MPQ student $\mathcal{Q}_{\theta}$. $X_i$ is the input sample. $c$ and $N$ denote the classes and the number of samples, respectively. Note that this process can be regarded as the distribution calibration for the student network, and one-hot label is not involved in training. 

\subsubsection{Entropy-aware Bin Regularization}\label{sec:ebr}
In previous quantization methods, real-valued weights are usually distributed uniformly. As a result, the quantized weight might collapse to a few quantization bins that close to zero. The information represented by the weights is heavily reduced during this process. According to the information theory, entropy should be preserved in quantization to retain the representation capacity of neural networks. 

In our method, after MPQ strategy is adopted, we propose \textbf{Entropy-aware Bin Regularization} to regularize weights in group of different bitwidth. The entropy carried by quantized weight with a specific bitwidth $b$ can be denoted as $\mathcal{H}_b(\bm W) = -\bm p_b(\bm w_i)\text{log}(\bm p_b(\bm w_i)), s.t. \sum_{i=1}^{2^b} \bm p_b(\bm w_i) = 1$, where $\bm p_b(\bm w_i)$ indicates the proportion of weights quantized to $i$-th quantization bin. 
We can derive that the entropy $\mathcal{H}_b(\bm W)$ is maximized when $\bm p_b^*(\bm w_i)=1/2^b$ for all $i \in \{1,\dots 2^b\}$. 
In light of the entropy analysis and inspired by previous research~\cite{li2019apot,yamamoto2021learnable}, we first normalize real-valued weights $\bm w_{l}^r$ of layer $l$ with bitwidth $b$ to $\bm w_{l}^{r*} = \frac{2^{(b-1)}}{2^b-1} \frac{|\bm w_{l}^r|}{||\bm w_{l}^r||_{1}}\bm w_{l}^r$. The corresponding quantized weights $\bm w_l^q$ are approximated to uniformly distribute in all quantization levels. Here, $|\bm w_l^r|$ is the number of entries in $\bm w_l^r$, and $||\bm w_l^r||_1$ computes the $L_1$-norm of $\bm w_l^r$.

Different from weight normalization in APOT~\cite{li2019apot} and LCQ~\cite{yamamoto2021learnable} which only consider the global weight distribution, we also want to minimize the quantization error in a specific quantization bin to reduce the information loss. Therefore, we regularize the distribution of the weights in a quantization bin to be a ``sharp'' Gaussian distribution (\textit{i.e.} a Dirac delta distribution ideally): its mean approaching quantization value and variance approaching zero. Motivated by these considerations, the Entropy-aware Bin Regularization is formulated as:
\vspace{-0.06in}
\begin{equation}
\label{eq:ebr}
    \mathcal{L}_{EBR}= \sum_{l=1}^{L} \sum_{n=1}^{2^{\bm b^{(l)}}} \mathcal{L}_{mse}(\overline{\bm w_{n,l}^r}, \bm w_{n,l}^q) + \mathcal{V}(\bm w_{n,l}^r), 
\end{equation}
where $\bm w_{n,l}^r, \bm w_{n,l}^q$ represent the real value and quantized value of weights in layer $l$ and quantization bin $n$. $\mathcal{L}_{mse}$ computes the mean square error, $\overline{(\cdot)}$ computes the mean, and $\mathcal{V}(\cdot)$ computes variance for all quantization bins with more than two elements. 

\subsection{Method Analysis and Discussion}
The complete algorithm of our proposed SDQ is summarized in Alg.~\ref{alg:SDQ}. Since our SDQ only performs stochastic quantization on weights, the precision of activations is unaffected throughout the quantization strategy generation phase. We freeze the activation bitwidth with a fixed value for the entire network during the training phase, and the activation is quantized in the same way as denoted by Eq.~\ref{equ:dorefa-quant}. Also note that our stochastic quantization is only performed on quantization strategy generation phase and is not applied during post-training or real inference.

While we only demonstrate how to use SDQ on layer-wise quantization. Our approach can easily be adapted to quantize models at many levels of granularity, such as block-wise, net-wise, and kernel-wise. However, the performance will be inhibited when using SDQ on coarse-grained granularity, such as blocks and networks, since the sensitivity difference between each layer will not be fully exploited. Additional training instability is expected when SDQ is used for fine-grained kernel-wise quantization because more quantization parameters are optimized alongside weight parameters during the MPQ strategy learning phase. Furthermore, current hardware accelerators cannot implement kernel-wise quantization, limiting the actual performance when deploying quantized models. Appendix~\ref{sec:granularity} contains further studies demonstrating that SDQ for layer-wise quantization produces the best results.

\begin{algorithm}[htb]
   \caption{Stochastic Differentiable Quantization}
   \label{alg:SDQ}
   {\bfseries Input:} the network with full precision weight $\{\bm{W}^r\}_{l=1}^L$, differentiable bitwidth parameters $\{\bm \beta\}_{l=1}^L$, bitwidth candidate set $\mathcal{B}$, maximum training epoch $\bm{E}_{\text{G}}$ (epoch for generating quantization strategy) and $\bm{E}_{\text{T}}$ (epoch for post-training), threshold $\bm \beta_{\text{t}}$ for decaying bitwidth \\ 
   {\bfseries Output:} quantized network with weights $\{\bm{W}^q\}_{l=1}^L$ and bitwidth allocations $\{\bm b^{(l)}\}_{l=1}^L$ 
\begin{algorithmic}[1]
   \STATE // Phase 1: Generating Quantization Strategy
   \FOR {$l = 1$ to $L$, $\bm b \in \mathcal{B}$}
   \STATE Initialize differentiable bitwidth parameters $\bm \beta_{l,b} = 1$
   \ENDFOR
   \FOR {epoch = 1 to $\bm{E}_{\text{G}}$}
   \STATE Quantize real-value weight $\bm{W}^r$ to $\bm{W}^q$ by Eq.~\ref{eq:prob_quant}
   \STATE Compute the task loss $\mathcal{L}_{task}$ and $\mathcal{L}_{QER}$ by Eq.~\ref{eq:qer}
   \STATE Compute gradient $\nabla \bm{W}^r$ and $\nabla \bm \beta$, update parameters
   \STATE Update $\{\bm b^{(l)}\}_{l=1}^L$ according to $\{\bm \beta\}_{l=1}^L$ and $\bm \beta_{\text{t}}$
   \ENDFOR
   \STATE Generate MPQ strategy $\{\bm b^{(l)}\}_{l=1}^L$ from DBPs $\{\bm \beta\}_{l=1}^L$
   \STATE // Phase 2: Post-training with Quantization Strategy
   \FOR {epoch = 1 to $\bm{E}_{\text{T}}$}
   \STATE Quantize real-value weights and activations by Eq.~\ref{equ:dorefa-quant}
   \STATE Compute the knowledge distillation loss $\mathcal{L}_{KD}$ and regularization $\mathcal{L}_{EBR}$ by Eq.~\ref{eq:kd} and Eq.~\ref{eq:ebr}
   \STATE Compute gradient $\nabla \bm{W}^r$, update parameters
   \ENDFOR
   \STATE \textbf{Return} MPQ network $\{\bm{W}^q\}_{l=1}^L$ with strategy $\{\bm b^{(l)}\}_{l=1}^L$
\end{algorithmic}
\end{algorithm}

\section{Experiments}
To evaluate the effectiveness of the proposed SDQ, we conduct experiments on the CIFAR and ImageNet-1K datasets. We first introduce the dataset, network, and training strategy in Sec.~\ref{sec:setting}, followed by the comparison with state-of-the-art quantization methods in Sec.~\ref{sec:sota}. We then analyze the effect and role of each proposed component of SDQ in Sec.~\ref{sec:ablation}. In Sec.~\ref{sec:vis}, we visualize the MPQ strategy and how our proposed SDQ improves the representation capacity of the quantized model. Hardware deployment experiments on different devices are designed to demonstrate the energy and time efficiency of our models in Sec.~\ref{sec:hardware}.

\subsection{Experimental Settings} \label{sec:setting}
\textbf{Dataset} The experiments are carried out on CIFAR-10 dataset~\cite{krizhevsky2009learning} and ImageNet-1K dataset~\cite{deng2009imagenet}. We only perform basic data augmentation in PyTorch~\cite{paszke2019pytorch}, which includes \textit{RandomResizedCrop} and \textit{RandomHorizontalFlip} during training, and single-crop operation during evaluation. 

\textbf{Network} We evaluate ResNet20 on CIFAR-10, and evaluate ResNet18 and MobileNetV2 on ImageNet-1K dataset. Due to the fact that the first and last layers are more sensitive to quantization perturbation compared to intermediate layers, we fix the bitwidth of them following previous work~\cite{yang2021fracbits}.

\textbf{Training detail} Following previous quantization methods~\cite{zhou2016dorefa, jung2019learning}, we adopt real-value pre-trained weights as initialization. We train and evaluate our models on various hardware platforms, including various NVIDIA GPUs and FPGA. Details of all hyper-parameters and training schemes are shown in Appendix~\ref{sec:details}.

\subsection{Comparison with State-of-the-Art Methods} \label{sec:sota}

Table~\ref{tab:cifar} compares our SDQ with existing methods for ResNet20 on CIFAR-10. We can see that our SDQ models significantly outperform the previous fixed precision quantization methods while also yielding better accuracy than other MPQ models with fewer average bitwidth. 

\begin{table}[ht]
\vspace{-0.2in}
\centering
\caption{Comparison with state-of-the-art mixed precision quantization methods (ResNet20 on CIFAR-10).}
\resizebox{0.48\textwidth}{!}{
\begin{tabular}{c|ccccccc}
\hline
\multirow{2}*{Method} & Bit-width & \multirow{2}*{mixed}  & \multicolumn{2}{c}{Accuracy(\%)} &\multirow{2}*{WCR}    \\ 
~                     &(W/A)      &~                      & Top-1 & FP Top-1                 & ~      \\
\hline
\hline
Dorefa\cite{zhou2016dorefa}     & 2/32            &  & 88.2        & 92.4           & 16$\times$          \\
PACT~\cite{choi2018pact}        & 2/32            &  & 89.7        & 92.4           & 16$\times$           \\
LQ-net\cite{zhang2018lqnet}     & 2/32            &  & 91.1        & 92.4           & 16$\times$          \\
\hline
TTQ~\cite{jain2019trained}      & 2.00/32        & \checkmark & 91.2        & 92.4           & 16$\times$ \\
Uhlich et al.~\cite{uhlich}     & 2.00/32        & \checkmark & 91.4        & 92.4           & 16$\times$ \\
BSQ~\cite{yang2020bsq}          & 2.08/32        & \checkmark & 91.9        & 92.6           & 15.4$\times$ \\
DDQ~\cite{zhaoyang2021DDQ}      & 2.00/32        & \checkmark & 91.6        & 92.4           & 16$\times$ \\
\hline
Ours                            & 1.93/32        & \checkmark &\textbf{92.1} & 92.4         & \textbf{16.6}$\times$ \\
\hline
\end{tabular}
}
\label{tab:cifar}
\end{table}

Table~\ref{tab:imagenet} shows the ImageNet-1K classification performance of our SDQ on ResNet18 and MobileNetV2. Since different methods adopt different full-precision (FP) models as initialization, we also report the corresponding Top-1 accuracy of the FP model for different methods.

Compared to the baseline FP model, our SDQ-trained models perform comparably or even better when quantized to low precision. For example, our ResNet18 achieves 71.7\% Top-1 accuracy when the average bitwdiths for weights and activations are 3.85 and 4 respectively, which has a 1.2\% absolute gain on the full-precision model. Our quantized MobileNetV2 is {\bf the first quantized model} with accuracy higher than FP initialization (72.0\% vs. 71.9\%), and average precision for weights and activations is lower than 4. 

With an optimal MPQ strategy and an effective training scheme, our SDQ outperforms previous quantization methods significantly under the same bitwidth configuration. To the best of our knowledge, the highest accuracy reported by previous uniform methods with weights and activation bitwidths smaller than 4 bits are 70.6\% on ResNet18 and 71.6\% on MobileNetV2 from FracBits-SAT~\cite{yang2021fracbits}. Our SDQ achieves an increase of 1.1\% on ResNet18 and 0.3\% on MobileNetV2 with a more compact bitwidth (3.85/4 vs. 4/4 on ResNet18, 3.79/4 vs. 4/4 on MobileNetV2). Note that SDQ even outperforms the state-of-the-art non-uniform quantization methods~\cite{li2019apot, chang2021rmsmp, zhaoyang2021DDQ}, proving the effectiveness and superiority of our method.

\definecolor{gray0}{gray}{.68}
\definecolor{gray1}{gray}{.78}
\definecolor{gray2}{gray}{.86}
\definecolor{gray3}{gray}{.91}
\definecolor{gray4}{gray}{.95}

\begin{table*}[ht]
\vspace{-0.1in}
\centering
\caption{Comparison with state-of-the-art mixed precision quantization methods (ResNet18 and MobileNetV2 on ImageNet-1K). Bitwidth (W/A) denotes the average bitwidth for weights and activation parameters. WCR represents the weight compression rate. BitOPs denotes the bit operations. For a filter $f$, the BitOPs is defined as $\text{BitOPs}(f)=b_wb_a|f|w_fh_f/s_f^2$, where $b_w$ and $b_a$ are the bitwidths for weights and activations, $|\cdot|$ denotes the cardinality of the filter, $w_f$,$h_f$,$s_f$ are the spatial width, height, and stride of the filter.}
\begin{threeparttable}[b]
\resizebox{\textwidth}{!}{
\begin{tabular}{cl|ccclclcc}
\hline
\multirow{2}*{Network}    & \multirow{2}*{Method} & Bit-width & \multirow{2}*{Mixed} & \multirow{2}*{Uniform}  & \multicolumn{2}{c}{Accuracy (\%)} &\multirow{2}*{WCR} &Model &\multirow{2}*{BitOPs (G)}\\ 
~                         &~                 &(W/A)      &~ &~                     & Top-1 & FP Top-1 & ~   & Size (MB) &~       \\
\hline
\hline
\multirow{16}*{ResNet18} &Dorefa\tnote{\dag}~\cite{zhou2016dorefa}   & 4/4  &  & \checkmark& 68.1 & 70.5  & 8$\times$ & 5.8 & 35.2\\
                          &PACT\tnote{\dag}~\cite{choi2018pact}      & 4/4  &  & \checkmark& 69.2 & 70.5  & 8$\times$ & 5.8 & 35.2\\
                          &LQ-net~\cite{zhang2018lqnet}  & 4/4  &  &           & 69.3 & 70.5  & 8$\times$ & 5.8 & 35.2\\
                          &APOT~\cite{li2019apot}                    & 4/4  &  &           & 70.7 & 70.5  & 8$\times$ & 5.8 & 34.7\\
                          \cline{2-10}
                          &DNAS\tnote{\dag}~\cite{wu2018mixed}       & -/-                     & \checkmark & \checkmark& 70.6           & 71.0 & 8$\times$   & 5.8 & 35.2\\
                          &HAQ~\cite{wang2019haq}                    & 4/32                    & \checkmark & \checkmark& 70.4           & 70.5 & 8$\times$   & 5.8 & 465\\
                          &EdMIPS~\cite{cai2020rethinking}           & 4/4                     & \checkmark & \checkmark& 68.0           & 70.2 & 8$\times$   & 5.8 & 34.7\\
                          &HAWQ-V3\tnote{\dag}~\cite{yao2021hawqv3}              & 4.8/7.5                 & \checkmark & \checkmark& 70.4           & 71.5 & 6.7$\times$ & 7.0 & 72.0\\
                          &Chen et al.~\cite{chen2021towards}        &\cellcolor{gray2}3.85/4  & \checkmark & \checkmark& \db{69.7}{0.1} & 69.8 & 8.3$\times$ & 5.6 & 33.4\\
                          &FracBits-SAT~\cite{yang2021fracbits}      & \cellcolor{gray1}4/4    & \checkmark & \checkmark& \ib{70.6}{0.4} & 70.2 & 8$\times$   & 5.8 & 34.7\\ 
                          \cline{2-10}
                          &Uhlich et al.~\cite{uhlich}               & 3.88/4  & \checkmark & & 70.1 & 70.3  & 8.3$\times$ & 5.6 &33.7 \\
                          &RMSMP~\cite{chang2021rmsmp}               & 4/4     & \checkmark & & 70.7 & 70.3  & 8$\times$   & 5.8 &34.7\\
                          &DDQ~\cite{zhaoyang2021DDQ}                & 4/4     & \checkmark & & 71.2 & 70.5  & 8$\times$   & 5.8 &34.7\\
                          \cline{2-10}
                          &\multirow{4}*{\textbf{Ours}}  &\cellcolor{gray1}3.85/8 & \checkmark & \checkmark& \textbf{\ib{72.1}{1.6}}  & 70.5   & \textbf{8.3}$\times$ & 5.6    &62.6  \\
                          &~                             &\cellcolor{gray2}3.85/4 & \checkmark& \checkmark & \textbf{\ib{71.7}{1.2}}  & 70.5   & \textbf{8.3}$\times$ & 5.6    &\textbf{33.4}  \\
                          &~                             &\cellcolor{gray3}3.85/3 & \checkmark & \checkmark& \textbf{\db{70.2}{0.3}}  & 70.5   & \textbf{8.3}$\times$ & 5.6    &\textbf{25.1}  \\
                          &~                             &\cellcolor{gray4}3.85/2 & \checkmark & \checkmark& \textbf{\db{69.1}{1.4}}  & 70.5   & \textbf{8.3}$\times$ & 5.6    &\textbf{16.7}   \\
\hline
\multirow{13}*{MobileNetV2}&Dorefa\tnote{\dag}~\cite{zhou2016dorefa}   & 4/4  &  & \checkmark & 61.8  & 71.9 &8$\times$ & 1.8 &7.42\\
                          &PACT\tnote{\dag}~\cite{choi2018pact}        & 4/4  &  & \checkmark & 61.4  & 71.9 &8$\times$ & 1.8 &7.42\\
                          &LQ-net~\cite{zhang2018lqnet}    & 4/4  &  &            & 64.4  & 71.9 &8$\times$ & 1.8 &7.42\\
                          &APOT~\cite{li2019apot}                      & 4/4  &  &            & 71.0  & 71.9 &8$\times$ & 1.8 &5.35\\
                          \cline{2-10}
                          &HAQ~\cite{wang2019haq}                      & 4/32                    & \checkmark & \checkmark& 71.5            & 71.9  & 8$\times$   & 1.8 &42.8\\
                          &HMQ~\cite{habi2020hmq}                      & 3.98/4                  & \checkmark & \checkmark& 70.9            & 71.9  & 8.1$\times$ & 1.7 &5.32\\
                          &Chen et al.~\cite{chen2021towards}          & \cellcolor{gray0}4.27/8 & \checkmark & \checkmark& \db{71.8}{0.1}  & 71.9  & 7.5$\times$ & 1.9 &5.32\\
                          &FracBits-SAT~\cite{yang2021fracbits}        &\cellcolor{gray2}4/4     & \checkmark & \checkmark& \db{71.6}{0.2}  & 71.8  & 8$\times$   & 1.8 &5.35 \\ 
                          \cline{2-10}
                          &Uhlich et al.~\cite{uhlich}                 & 3.75/4      & \checkmark & & 69.8 & 70.2 & 8.5$\times$ & 1.6 &5.01\\
                          &RMSMP~\cite{chang2021rmsmp}                 & 4/4         & \checkmark & & 69.0 & 71.9 & 8$\times$   & 1.8 &5.35 \\
                          &DDQ~\cite{zhaoyang2021DDQ}                  & 4/4         & \checkmark & & 71.8 & 71.9 & 8$\times$   & 1.8 &5.35 \\
                          \cline{2-10}
                          &\multirow{2}*{\textbf{Ours}}  &\cellcolor{gray1}3.79/8 & \checkmark & \checkmark &\textbf{\ib{72.9}{1.0}}  & 71.9 & \textbf{8.4}$\times$ & 1.6 &10.1  \\
                          &                              &\cellcolor{gray2}3.79/4 & \checkmark & \checkmark &\textbf{\ib{72.0}{0.1}}  & 71.9 & \textbf{8.4}$\times$ & 1.6 &\textbf{5.07}  \\
                          
\hline
\end{tabular}}
\begin{tablenotes}
     \item[\dag] re-implementation for fair comparisons under the same backbone architectures
\end{tablenotes}
\end{threeparttable}
\label{tab:imagenet}
\vspace{-0.25in}
\end{table*}

\subsection{Ablation Study} \label{sec:ablation}
We further conduct ablation studies to prove the contribution of different components to our model's performance, including SDQ quantization strategy generation, knowledge distillation, and Entropy-aware Bin Regularization. Table~\ref{tab:strategy} compares different quantization strategy generation schemes impartially on the same initialization and training. The performance of the quantized model with the quantization strategy generated by our SDQ is close to previous MPQ models, while the average bitwidths of our strategy are lower. This comparison reflects that our strategy generation contributes to the improvement of accuracy.

\begin{table}[H]
\vspace{-0.2in}
\centering
\caption{Comparison with state-of-the-art mixed precision quantization strategy generation methods under same training and initialization (MobileNetV2 on ImageNet-1K).}
\resizebox{0.45\textwidth}{!}{
\begin{tabular}{c|ccccccc}
\hline
\multirow{2}*{MPQ Strategy} & Bit-width & \multicolumn{2}{c}{Accuracy(\%)} \\ 
~                     &(W/A)      & Top-1 & Top-5                 \\
\hline
\hline
HMQ~\cite{habi2020hmq}         & 3.98/4      & 71.7        & 90.1       \\
FracBits~\cite{yang2021fracbits}    & 4/4         & 72.0        & 90.4       \\
Our strategy                        & \textbf{3.79/4}      & \textbf{72.0}        & \textbf{90.5}       \\
\hline
\end{tabular}
}
\vspace{-0.25in}
\label{tab:strategy}
\end{table}

The comparison of different weight regularization methods is shown in Table~\ref{table:ablation_w}. The experiments are conducted on MPQ ResNet18 with average bitwidths of 3.85 and 2 for weights and activations. The insights of our proposed EBR are to penalize the global distribution of weights to be over centralized, and penalize the local distribution of weights in different quantization bins to be too dispersed. The effect of it is also shown in Fig.~\ref{Figure:compare-ebr}. Compared to other weights regularization methods which only consider the global distribution, our EBR significantly improves the quantization robustness of MPQ models. 

\begin{table}[H]
\vspace{-0.2in}
\begin{center}
\caption{Comparison of our Entropy-aware Bin Regularization (EBR) and other weight regularization methods for our mixed precision quantized ResNet18 on ImageNet-1K.}
\vspace{-0.1in}
\resizebox{0.49\textwidth}{!}{
\begin{tabular}{c|ccc}
\hline
\multicolumn{2}{c}{Method} & Top-1 Acc & Top-5 Acc\\
\hline
\hline
\multicolumn{2}{c}{Baseline} & 67.6 & 87.6 \\ 
\multicolumn{2}{c}{Weight Norm~\cite{salimans2016weight}} & 66.6 & 86.7\\ 
\multicolumn{2}{c}{KURE~\cite{shkolnik2020robust}} & 68.5 & 88.4\\ 
\hline
 & $\lambda_E=0.01$ &68.6 & 88.4\\
EBR & $\lambda_E=0.1$ &\textbf{69.1} & \textbf{88.5}\\
 & $\lambda_E=1$ & 68.9 & 88.4 \\
\hline
\end{tabular}}
\label{table:ablation_w}
\end{center}
\vspace{-0.25in}
\end{table}

How the knowledge distillation contributes to the training of MPQ models is shown in Table~\ref{table:ablation_kd}. Learning from FP models helps minimize the performance gap of quantization models. When the performance of FP teacher models is better, the performance of student models also improves.

\begin{table}[H]
\vspace{-0.2in}
\begin{center}
\caption{Comparison of different teacher models of knowledge distillation for our mixed precision quantized ResNet18 on ImageNet-1K. One-hot label with CE loss is used for ``w/o KD'' in experiment.}
\resizebox{0.45\textwidth}{!}{
\begin{tabular}{cccccc}
\hline
 Method & Teacher & Top-1 Acc & Top-5 Acc\\
\hline
\hline
Ours w/o KD & Ground Truth & 70.5 & 89.5 \\ 
\hline
     & ResNet34  &70.7 & 89.7\\
Ours & ResNet50  &71.1 & 89.9\\
     & ResNet101 & \textbf{71.7} & \textbf{90.2}\\
\hline
\end{tabular}}
\label{table:ablation_kd}
\end{center}
\vspace{-0.2in}
\end{table}

\subsection{Visualization and Analysis} \label{sec:vis}
\begin{figure*}[t]
\vspace{-0.1in}
\centering
\subfigure[ResNet18 (3.85bit)]{%
\label{fig:resnet18-mbit}
\includegraphics[width=0.26\textwidth]{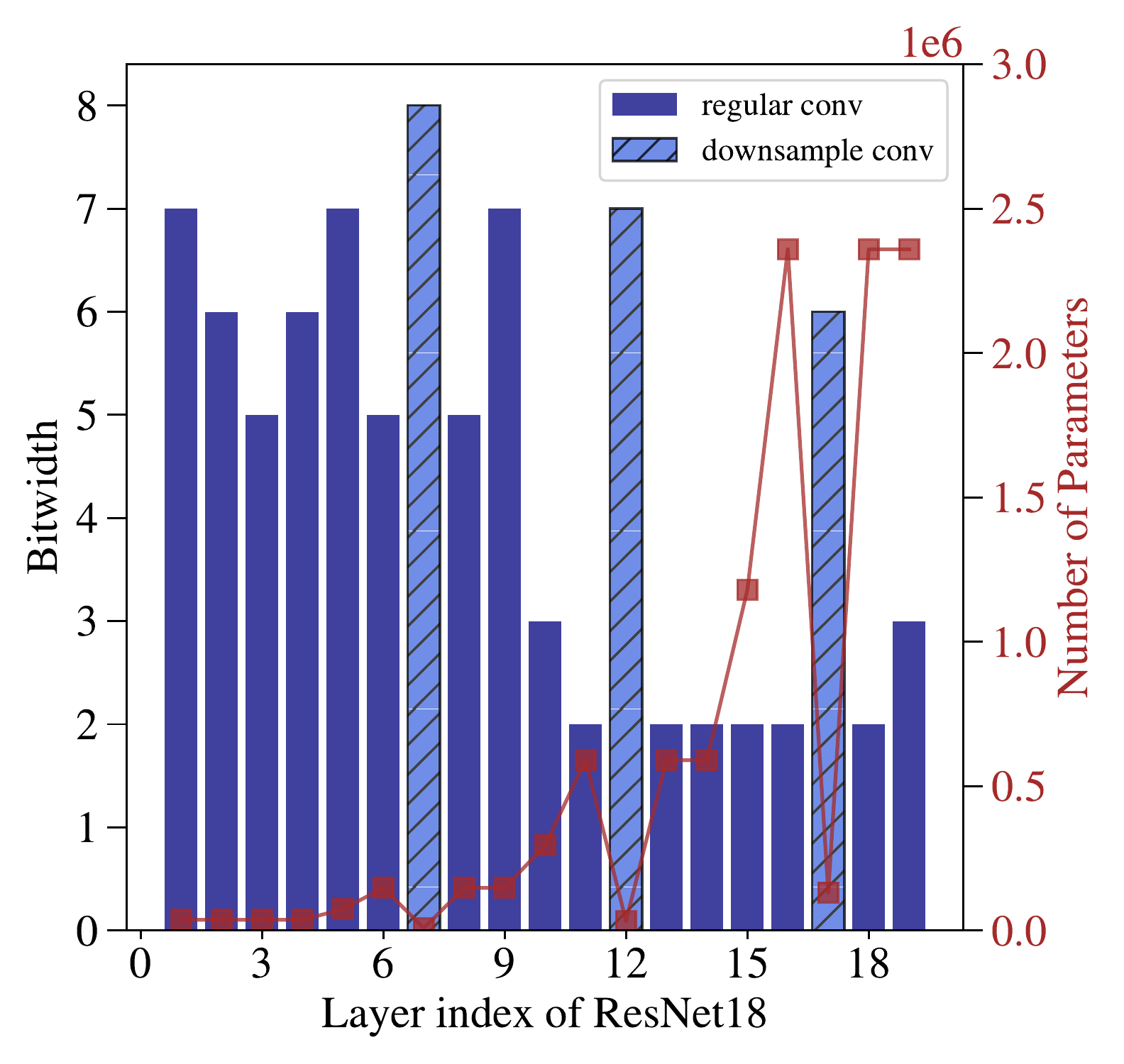}}%
\subfigure[MobileNetV2 (3.79bit, The first 3$\times$3 conv layer and last fc layer is not visualized here)]{%
\label{fig:mbnetv2-mbit}
\includegraphics[width=0.7\textwidth]{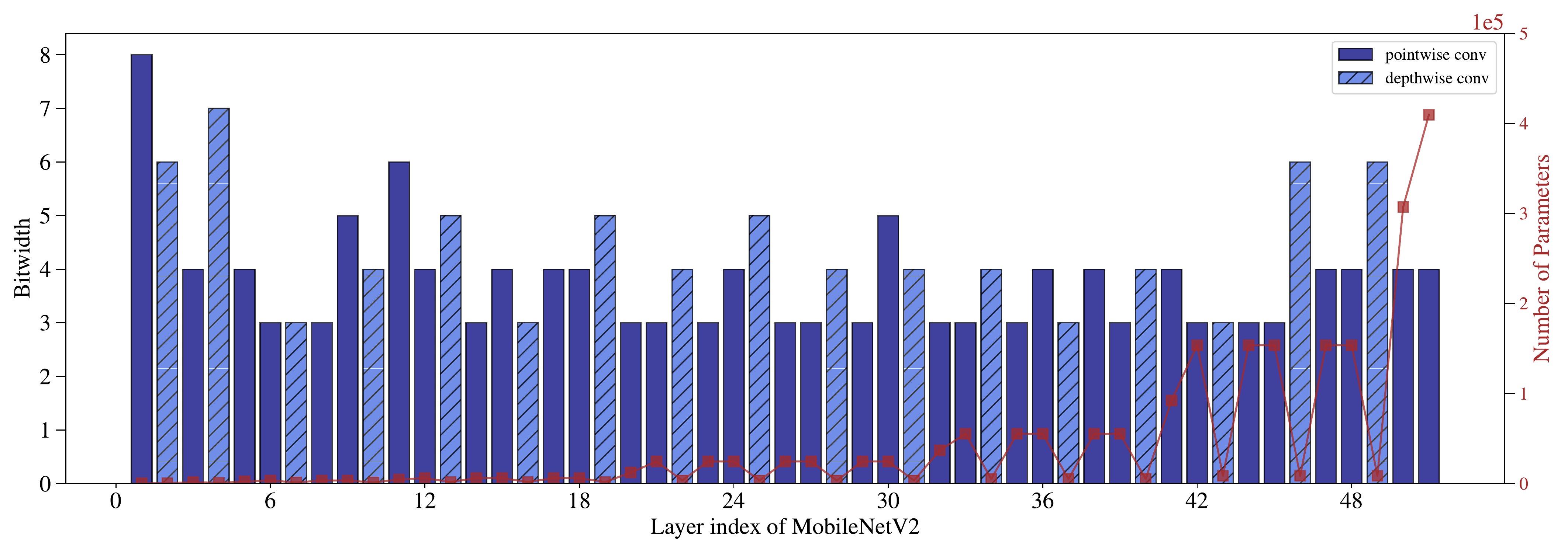}
}%
\vspace{-0.1in}
\caption{Mixed Precision Quantization strategy generated by our SDQ for ResNet18 and MobileNetV2 on ImageNet-1K dataset.}
\vspace{-0.1in}
\end{figure*}

\begin{figure}[H]
\vspace{-0.1in}
    \centering 
	\includegraphics[width=0.47\textwidth]{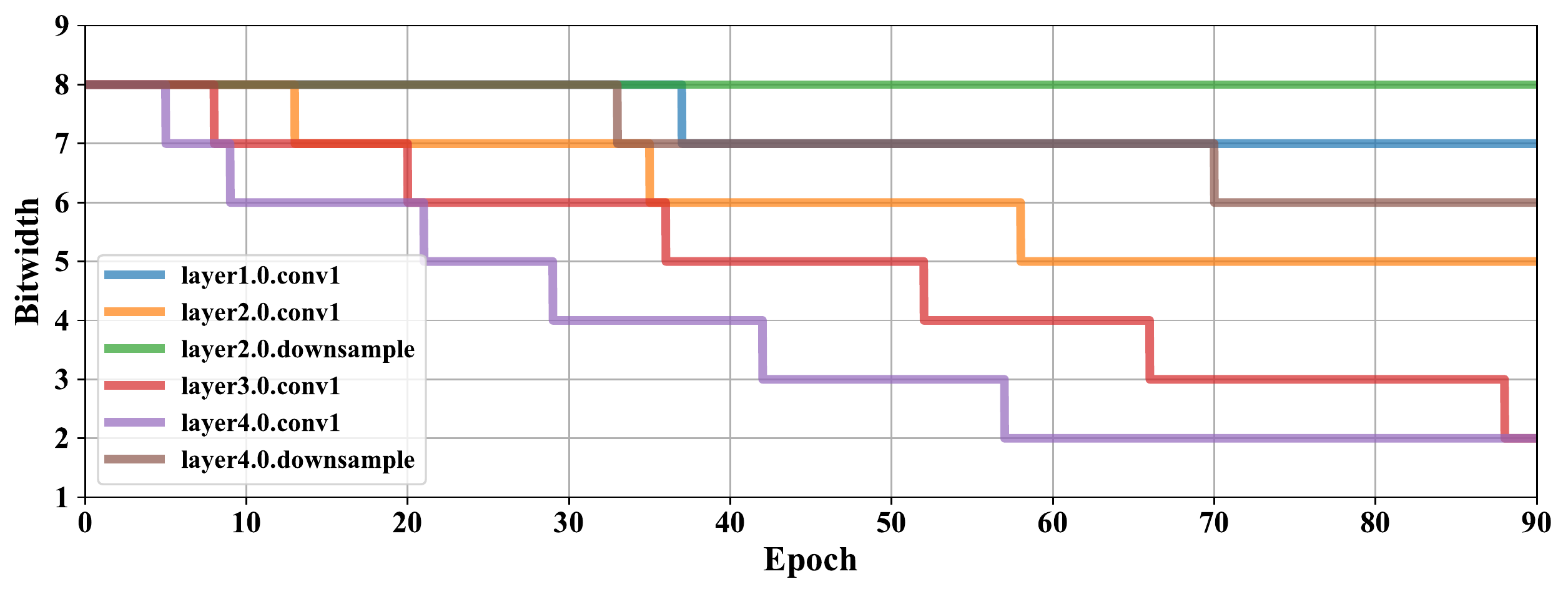}
	\vspace{-0.1in}
	\caption{Bitwidth assignment of selected layers in ResNet18 during quantization strategy generation.}
	\label{Figure:bitw-change}
	\vspace{-0.15in}
\end{figure}

To intuitively demonstrate the bitwidth assignment generated by our SDQ, the quantization strategy of weights in different layers of ResNet18 and MobileNetV2 is visualized in Fig.~\ref{fig:resnet18-mbit} and Fig.~\ref{fig:mbnetv2-mbit}, respectively. Our SDQ learns more bits for those layers with fewer parameters (\textit{downsample conv} layers in ResNet18 and \textit{depthwise conv} layers in MobileNetV2), which proves that more redundancy is expected for those layers. As layers with the same number of parameters can have different optimal bitwidths, it can be proved that there is no simple heuristics to allocate bitwidth. We can also see from the bitwidth assignment that the first few layers and the last few layers have higher bitwidth. Fig.~\ref{Figure:bitw-change} depicts how bitwidth assignment evolves during the quantization strategy generation phase. While the decay of bitwidth is slow for \textit{downsample conv} layers, the regular \textit{conv} layers decay quickly at high precision and slow down when it reaches low precision.

\begin{figure}[h]
\vspace{-0.1in}
\centering
\subfigure[Baseline (2 bit)]{%
\label{fig:tsne-baseline}%
\includegraphics[width=0.2\textwidth]{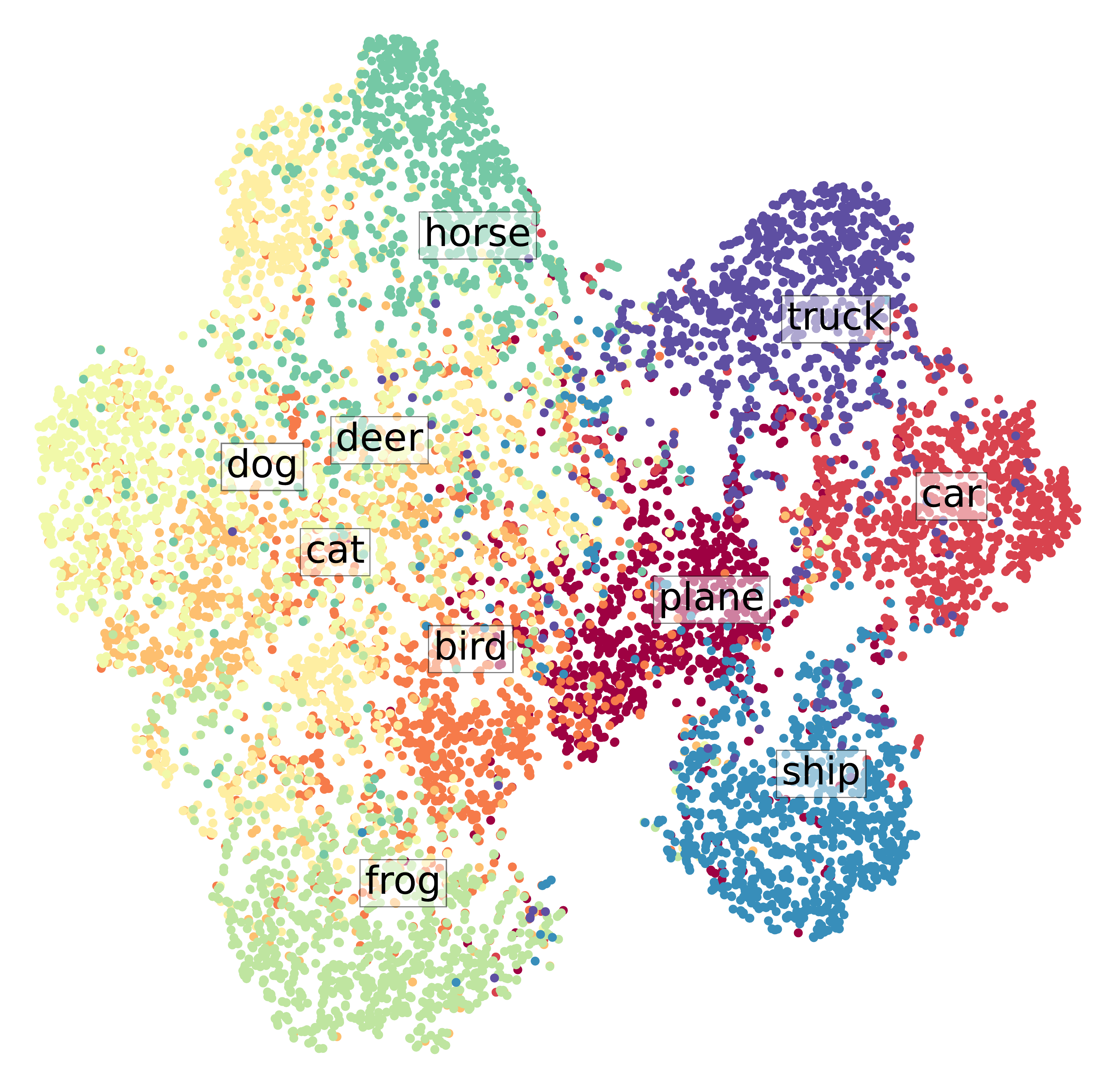}}%
\subfigure[Mixed Ours (1.93 bit)]{%
\label{fig:tsne-mixed}%
\includegraphics[width=0.2\textwidth]{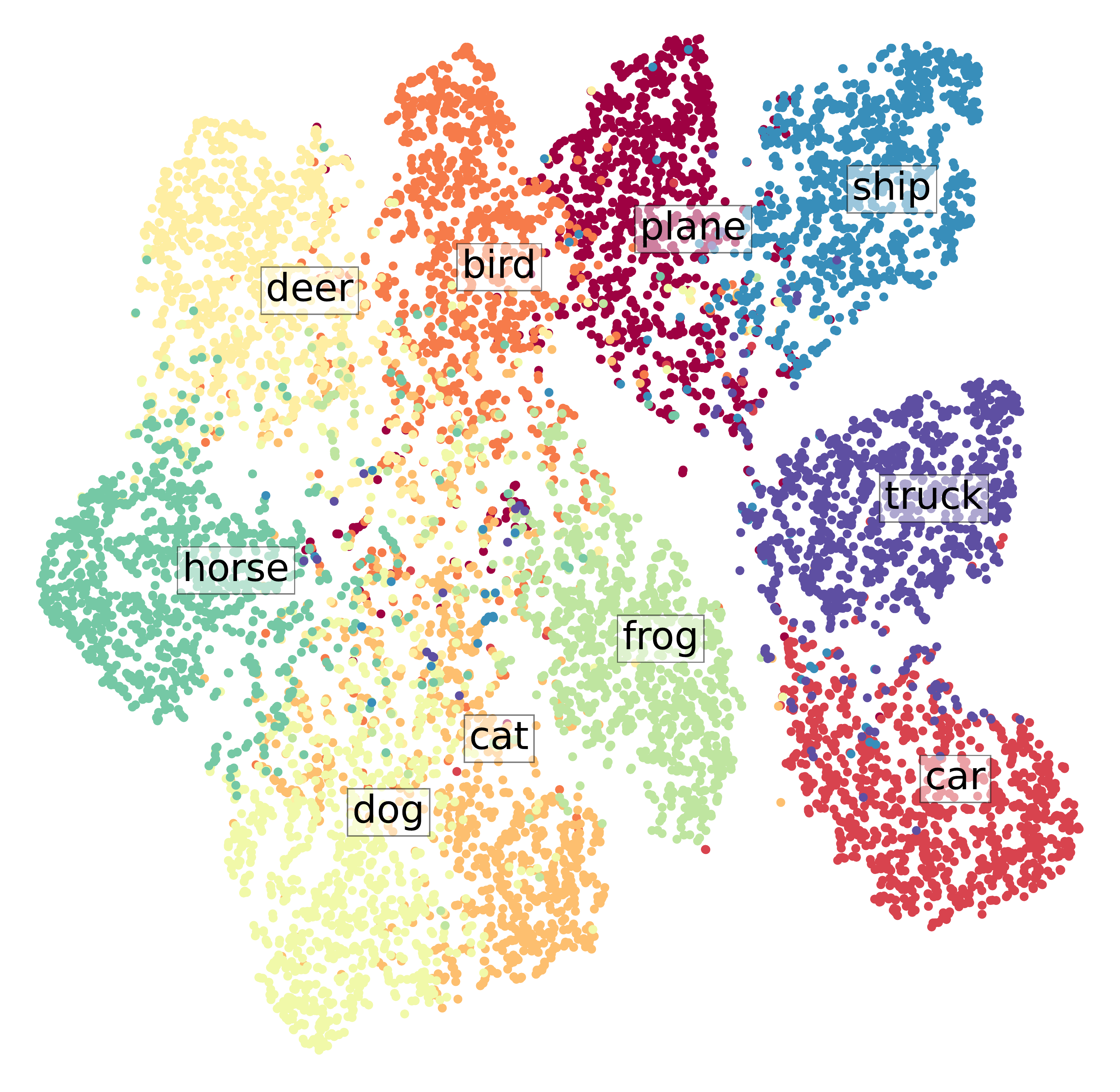}}%
\vspace{-0.1in}
\caption{Comparison of feature embedding visualization using t-SNE on CIFAR-10 evaluation set. The feature embedding are extracted from the output of last conv layer of ResNet20. The performance of these two models are reported in Table~\ref{tab:cifar}.}
\end{figure}

Compared to uniform quantization, the MPQ network performs better with fewer bits since the MPQ network fully utilizes the difference of layers' sensitivity to quantization. Fig.~\ref{fig:tsne-baseline} and Fig.~\ref{fig:tsne-mixed} visualize the feature embedding of last conv layer in ResNet20. As expected, the clusters of dog and deer are mixed up together in the baseline model with 2-bit uniform quantization. While the model of SDQ can split those classes with more separable representations. 

\begin{figure}[h]
\vspace{-0.1in}
\centering
\subfigure[Weights w/o EBR]{%
\label{fig:ebrweight}%
\includegraphics[width=0.16\textwidth]{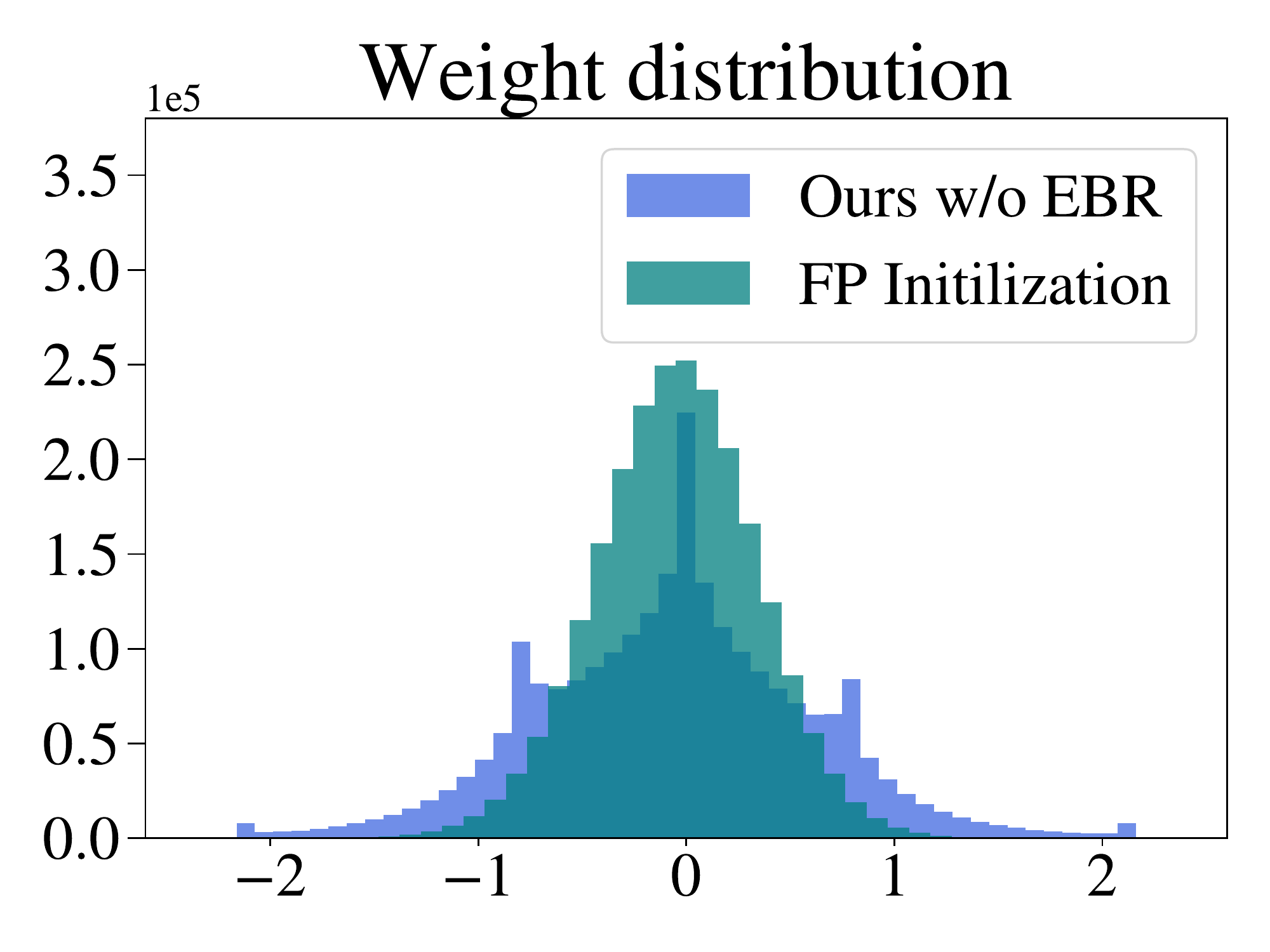}}%
\subfigure[Weights w/ EBR]{%
\label{fig:noebrquant}%
\includegraphics[width=0.16\textwidth]{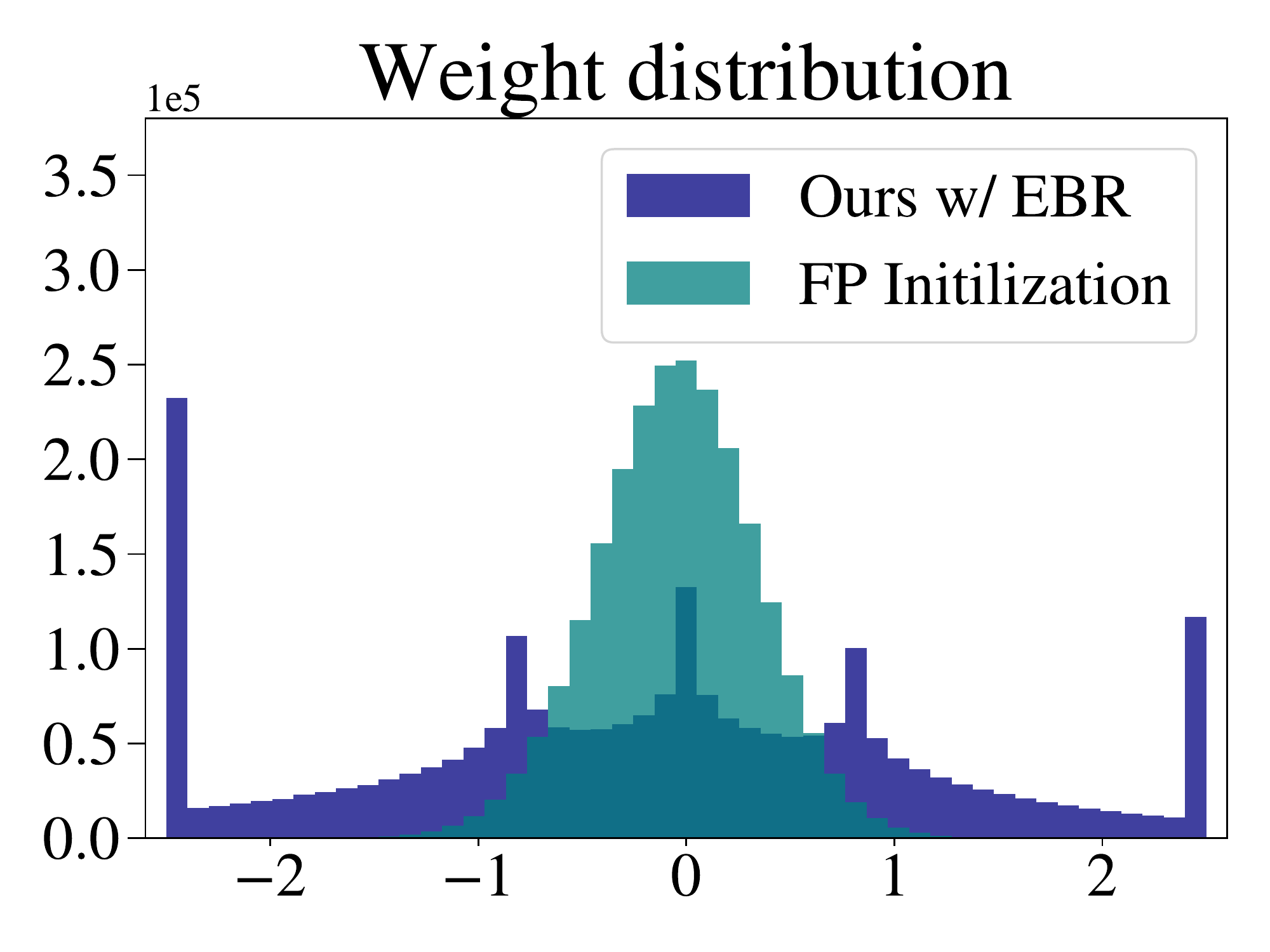}}%
\subfigure[Quantization]{%
\label{fig:comparequant}%
\includegraphics[width=0.16\textwidth]{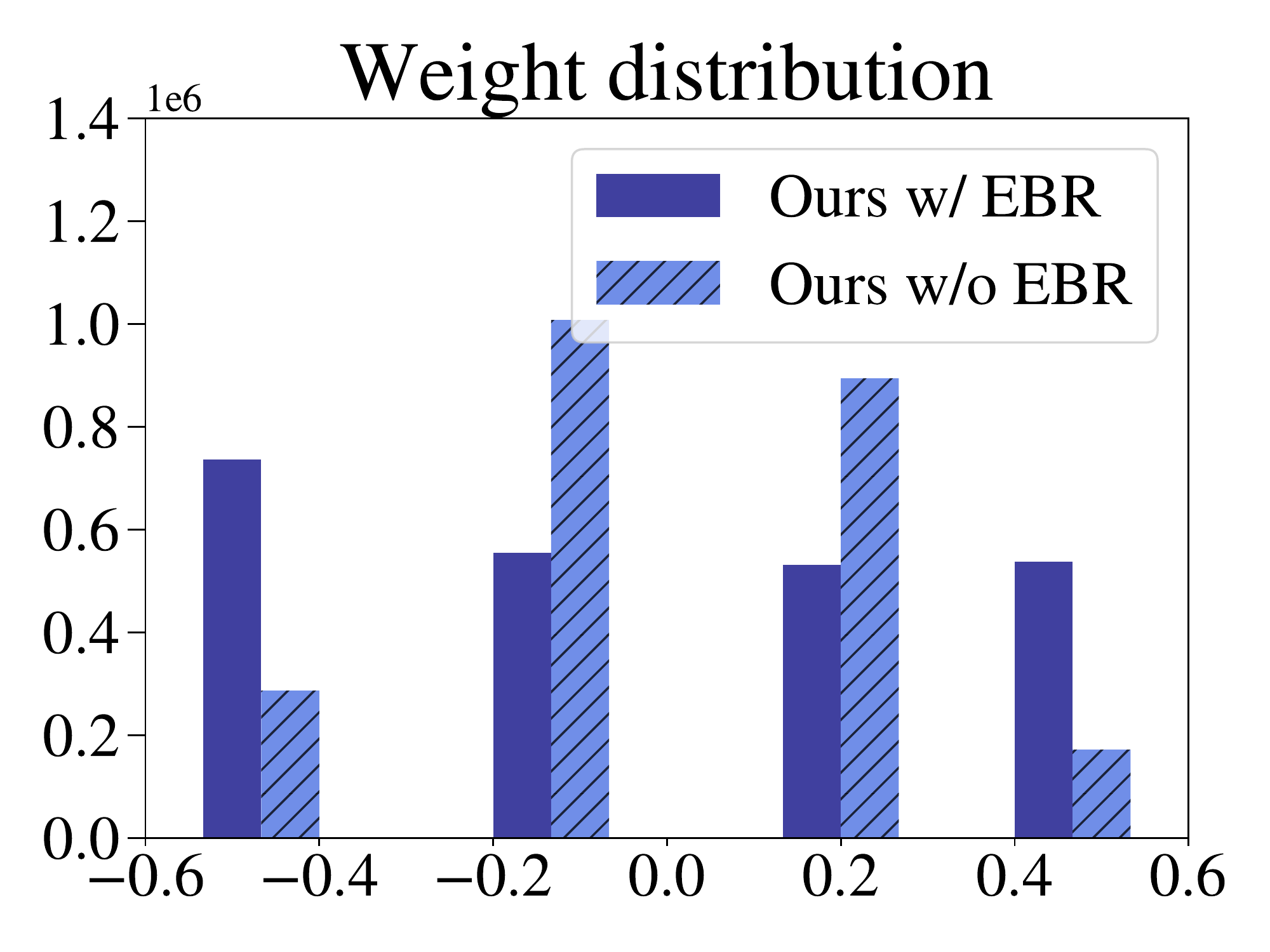}}%
\vspace{-0.1in}
\caption{Histogram of the weight distribution and quantization distribution. \ref{fig:ebrweight} and \ref{fig:noebrquant} are the histogram of weights of \textit{layer4.1.conv2} with 2-bit precision in our quantized ResNet18 model with and without EBR respectively. \ref{fig:comparequant} compares the quantization results with and without EBR.}
\label{Figure:compare-ebr}
\vspace{-0.15in}
\end{figure}

We proposed Entropy-aware Bin Regularization to preserve the entropy and minimize the quantization error. Fig.~\ref{Figure:compare-ebr} shows the histogram of weight distribution in full-precision and 2-bit quantization. 
We can observe that the weights in our model are more centralized locally to the quantization point in each quantization bin. Further, the portion of weights quantized to each bin in our model is closer compared to baseline, which preserves the entropy globally.

\subsection{Hardware Efficiency on Accelerator} \label{sec:hardware}
We further conduct hardware experiments to evaluate the efficiency of our MPQ model. In Table~\ref{table:hardware}, we evaluate our model on the accelerator that supports mixed precision arithmetic operation: Bit Fusion~\cite{sharma2018bit}. Bit Fusion only supports multiplications of power-of-two bits (\textit{i.e.}~2, 4, 8, 16 bits), so there is a certain performance gap between theoretical compression and real compression. Our ResNet18 model with an average weight bitwidth of 3.85 achieves much better accuracy than the ResNet18 model with a weight bitwidth of 4, and surpasses its time cost and energy efficiency. 

\begin{table}[H]
\vspace{-0.1in}
\begin{center}
\caption{Comparison of latency and energy consumption of quantized ResNet18 model on Bit Fusion.}
\resizebox{0.48\textwidth}{!}{
\begin{tabular}{cccccc}
\hline
Method & Bitwidth (W/A) & Top-1 Acc &  Latency & Energy\\
\hline
\hline
\rowcolor{grayDark}
Dorefa& 4/8   & 69.8         & 48.99 ms  &93.34 mJ\\ 
Ours&  3.85/8 &\textbf{72.1} & 46.18 ms  &90.18 mJ\\
\rowcolor{grayDark}
Dorefa& 4/4   & 67.1         & 34.61 ms  &64.16 mJ\\ 
Ours&  3.85/4 &\textbf{71.1} & 32.84 ms  &60.49 mJ\\
\rowcolor{grayDark}
Dorefa& 4/2   & 63.6         & 30.77 ms  &54.08 mJ\\ 
Ours&  3.85/2 &\textbf{69.1} & 28.18 ms  &49.05 mJ\\
\hline
\end{tabular}}
\label{table:hardware}
\end{center}
\end{table}

\subsection{COCO Detection Hardware Experiment}

We also evaluate our SDQ on object detection task which is more challenging. We implement the mixed-precision compact YOLOv4-tiny~\cite{bochkovskiy2020yolov4} with our SDQ on COCO detection dataset~\cite{lin2014coco}, which consists of 80 object categories. The training set contains 115k images (\texttt{trainval35k}), and the validation set has 5k images (\texttt{minival}). Different from previous research~\cite{li2019fully, wang2020bidet} that apply quantized models on Faster R-CNN~\cite{ren2015faster} or RetinaNet~\cite{lin2017focal} using ResNet and MobileNet as backbones, YOLOv4-tiny is already compact and sensitive to the quantization. 
Our quantization strategy generation and training phase are conducted on \texttt{trainval35k} partition. During the training phase, percentile activation calibration~\cite{li2019fully} is used to discard outlier activation values. Since our target deployment platform only supports power-of-two bitwidth, we set the bitwidth candidate $\mathcal{B}=\{1,2,4,8\}$.
The input size for the object detector is $416 \times 416$. 
Following the standard COCO evaluation metric, we report the average precision (AP) for different IoU thresholds on the \texttt{minival} partition. 
The results, including latency and energy consumption when deploying on the FPGA system are shown in Table~\ref{tab:coco}.

\begin{table*}[t]
\vspace{-0.1in}
\centering
\caption{Performance comparison on COCO detection benchmark. }
\resizebox{\textwidth}{!}{
\begin{tabular}{c|clccccccccc}
\hline
\multirow{2}*{Method}&  Bitwidth & \multirow{2}*{Hardware} & \multirow{2}*{AP} & \multirow{2}*{$\text{AP}_{50}$} & \multirow{2}*{$\text{AP}_{75}$}  &\multirow{2}*{$\text{AP}_{S}$}  &\multirow{2}*{$\text{AP}_{M}$} &\multirow{2}*{$\text{AP}_{L}$} & \multirow{2}*{Latency} & \multirow{2}*{Energy} & \multirow{2}*{FPS} \\ 
~                    &  (W/A)    & ~ & ~   & ~   & ~  & ~  & ~   \\
\hline
YOLOv4-tiny~\cite{bochkovskiy2020yolov4}   & 32/32 & GPU  & 20.4 & 38.7 & 19.5 & 7.4 & 24.1 & 27.1    &-&-&-\\
\hline
\multirow{2}*{Dorefa~\cite{zhou2016dorefa}}& 8/8  & FPGA    & 16.1 & 32.3 & 14.5 & 4.3 & 17.3 & 24.7  &34.18ms & 268.3mJ &29\\
~                                          & 4/4  & FPGA    & 15.4 & 29.2 & 12.5 & 3.7 & 16.9 & 23.3  &18.64ms & 146.3mJ &53\\
\hline
 Ours + Dorefa                          & 3.88/4 & FPGA  & 15.9 & 31.1 & 14.7 & 4.2 & 17.6 & 24.2  &21.28ms & 167.1mJ &47\\
\hline
\end{tabular}
}
\label{tab:coco}
\vspace{-0.1in}
\end{table*}

From the table we can see that our SDQ improves the efficiency of object detector without incurring significant mAP degradation compared to the baseline quantization approach (8/8 bitwidths), while our detector with lower bitwidths (3.88/4) outperforms the 4-bit quantized model favorably. According to the FPGA deployment statistics, our SDQ demonstrates good hardware affinity yielding latency and energy consumption close to the 4-bit uniform single-precision model. Overall, mixed-precision YOLOv4-tiny trained with our SDQ can achieve similar performance of 8-bit quantized model while retaining the hardware efficiency of 4-bit quantized model.

\subsubsection{FPGA System Setting}
\begin{figure}[H]
\vspace{-0.25in}
\centering
\includegraphics[width=0.4\textwidth]{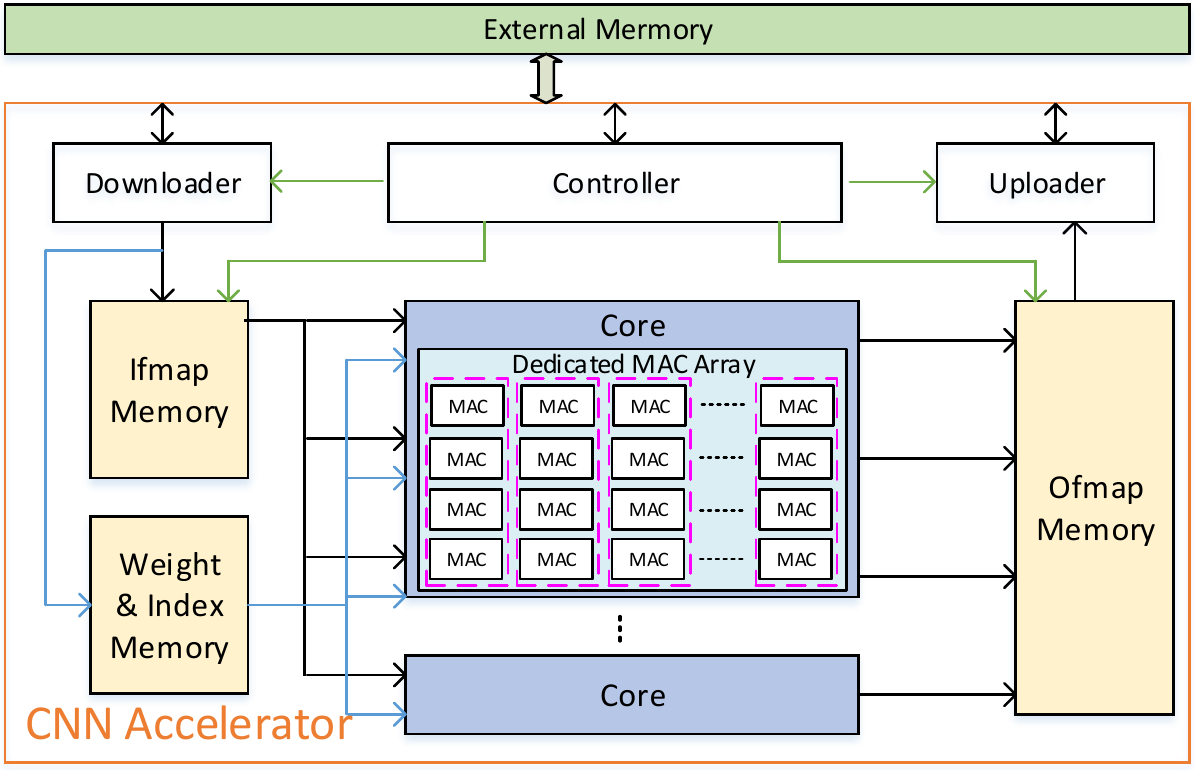}
\vspace{-0.1in}
\caption{System architecture schematic. The parameters of our hardware experiment platform are: MAC array-4 rows $\times$ 16 columns, Frequency-200MHz, Number of Cores-8.}
\label{Figure:hardware-arch}
\vspace{-0.2in}
\end{figure}

The system is implemented on the Xilinx U50 FPGA platform and consumes 259688 LUTs and 210.5 BRAM. As shown in Fig.~\ref{Figure:hardware-arch}, a CNN accelerator system is comprised of multiple cores, controller, on-chip memory, downloader, and uploader. Each core consists of 4$\times$16 dedicated array of INT-8 multiply-and-accumulate (MAC) processing elements. All cores share the same on-chip memory, which stores the input feature map (ifmap), weight, index, and output feature map (ofmap). The index is used for weight sparse to scratch the input activation. The controller module uses instructions to take charge of convolution computation, data download, and upload. 

\section{Conclusion}
We have presented \textbf{S}tochastic \textbf{D}ifferentiable \textbf{Q}uantization \textbf{(SDQ)}, a learning based mixed precision quantization framework that can optimize the optimal MPQ strategy automatically. 
In contrast to previous MPQ methods, SDQ introduces a set of Differentiable Bitwidth Parameters (DBPs) that follows a stochastic quantization scheme to make DBPs differentiable with a stabler and smoother optimization process. Since SDQ is optimized on a global searching space, the learned MPQ strategy is usually better than other competitors. 
Moreover, Quantization Error Regularization (QER) and Entropy-aware Bin Regularization (EBR) are integrated into the MPQ learning and post-training stages to minimize the quantization error. We demonstrate the superiority of our SDQ with comprehensive experiments on different hardware platforms such as GPUs and FPGA across various networks. 
SDQ achieves state-of-the-art accuracy over previous quantization methods with fewer average bits. Extensive hardware experiments prove the superior efficiency of our models deployed on mixed-precision accelerators. 

\vspace{-0.1in}
\section*{Acknowledgements}
\vspace{-0.05in}
This research was partially supported by ACCESS - AI Chip Center for Emerging Smart Systems, sponsored by InnoHK funding, Hong Kong SAR. 

\bibliography{example_paper}
\bibliographystyle{icml2022}

\clearpage
\appendix
\onecolumn

\begin{center}
     \section*{Appendix}
\end{center}

This appendix includes additional analysis, implementation details, and extended experimental analysis not included in the main text due to space limitation. These contents are organized in separate sections as follows:
\begin{itemize}
    \item Sec.~\ref{sec:qer-analysis} analyzes how quantization error varies within different layers using mixed precision and how we propose a weighting coefficient to balance between the precision.
    \item Sec.~\ref{sec:granularity} elaborates the performance of MPQ in different granularities of networks and the reasons why layer-wise quantization is optimal.
    \item Sec.~\ref{sec:details} includes the training details and experiment settings, such as the hyper-parameters, training dynamics, and detailed compression of bitwidth assignment as shown in Table~\ref{tab:hyperpara}.
\end{itemize}

\section{Quantization Error Analysis} \label{sec:qer-analysis}

Quantization error $\bm \Omega$ is defined as $L_2$ distance $||\bm x^q-\bm x^r||_2$ between quantized value $\bm x^q$ and real value $\bm x^r$. In uniform quantization, $\bm \Omega_u$ depends on the characteristics of layers which includes parameter amounts, learned weight distribution, and bitwidth assignment. Among these factors, the most important one is the bitwidth of the layers. As shown in Table~\ref{table:quanterror}, the squared quantization error $\bm \Omega_u^2$ grows exponentially with the bitwidth.

\begin{table}[H]
\begin{center}
\caption{Comparison of squared quantization error $\bm \Omega_u^2$ for various bitwidths of different layers in ResNet20.}
\resizebox{0.6\textwidth}{!}{
\begin{tabular}{c|c|ccccc}
\hline
Layer & Parameter Size & 8-bit & 6-bit &  4-bit & 3-bit & 2-bit \\
\hline
\hline
Layer1.2.conv1  & 2.3e4 & 0.03 & 0.52 & 9.25 & 41.9 & 191\\
Layer2.1.conv1  & 9.2e4 & 0.05 & 0.81 & 14.0 & 65.1 & 309\\
Layer3.1.conv1  & 3.7e5 & 0.29 & 4.83 & 84.8 & 392  & 2056\\
\hline
\end{tabular}}
\label{table:quanterror}
\end{center}
\end{table}

Uniform quantization defined in Eq.~\ref{equ:dorefa} and Eq.~\ref{equ:dorefa-quant} linearly maps full-precision weights $\bm w^r$ into quantized representations $\bm w^q$. To remove the outlier of weight distribution, clamp operation is used to restrict the real value in the range $[\bm w_l, \bm w_u]$. The quantized output $\bm w^q$ from uniform quantization scheme at $\bm b$-bit with clamp can be defined as: 
 
\begin{equation} \label{eq:uniform-quant-scheme}
\begin{aligned}
\bm w^q =  \bm Q_{u} (\bm w^r; b, \bm w_{l}, \bm w_{u})  =  \bm s \times \text{round}[\text{clamp} (\bm w^r; \bm w_{l}, \bm w_{u})/\bm s], 
\end{aligned}
\end{equation}

where $[\bm w_l, \bm w_u]$ is the quantization range, $\bm \Delta=\bm w_{u}-\bm w_{l}$ is the range length, $\bm s=\frac{\Delta}{\bm N-1}$ is the scaling factor, $\bm N=2^b$ is the total number of quantization bins. Previous work~\cite{you2010audio} has proved that the expected quantization error squared for the $b$-bit uniform quantizer is denoted as:
\begin{equation} \label{eq:uni-quant-err}
\mathbb{\bm E} (\bm \Omega_{u}^2; b, \bm w_{l}, \bm w_{u}) = \frac{\bm s^2}{12} = \bm C(b) \bm \Delta^2, 
\end{equation}
where $\bm C(b) = \frac{1}{12 (2^b-1)^2}$ for the uniform distributions. This is the reason that we use $\bm \lambda_b=(2^{b}-1)^2$ in Eq.~\ref{eq:qer} to balance between different bitwidths. In our stochastic quantization scheme, we assume the same quantization range $[\bm w_l, \bm w_u]$ is used for neighboring quantization levels. The expectation of quantization error at $b_i$ bitwidth DBP $\bm \beta$ can be computed as:
\begin{equation} \label{eq:sdq-err}
\begin{aligned}
\mathbb{E} (\bm \Omega_{s}^2; b_i) = \bm \beta_{b_i} \mathbb{E} (\bm \Omega_{u}^2; b_i) + (1-\bm \beta_{b_i}) \mathbb{E} (\bm \Omega_{u}^2; b_{i-1}) = [\bm \beta_{b_i} \bm C(b_i) + (1-\bm \beta_{b_i}) \bm C(b_{i-1})] \bm \Delta^2, 
\end{aligned}
\end{equation}
where $\bm C(b_i) = \frac{1}{12 (2^{b_i}-1)^2}$. Accordingly, the coefficient $\bm \lambda_b=(2^{b_i}-1)^2$ can still be applied to balance the regularization term at neighboring quantization levels while amplifying the quantization error for lower bit $b_{i-1}$ by $[(2^{b_i}-1)/(2^{b_{i-1}}-1)]^2$.

\section{SDQ on Different Granularities} \label{sec:granularity}

Our method uses the differentiability of DBPs to learn the bitwidth automatically. The advantage of the proposed SDQ is that DBPs can be inserted into different network granularities flexibly. Table~\ref{table:granularity} compares the model performance, epochs for MPQ strategy generation, and optimization time per epoch using ResNet18 on different granularities. 

\begin{table}[H]
\begin{center}
\caption{Comparison of Top-1 accuracy, total epochs for strategy generation, and time per epoch of different granularities with ResNet18 on ImageNet-1K.}
\begin{tabular}{c|cccccc}
\hline
Granularity & Bit-width(W/A) & Top-1 Acc & Epochs & Time/epoch \\
\hline
\hline
Net     & 4/4    & 68.7 & 30 & 47min \\
Block   & 3.87/4 & 71.2 & 60 & 47min \\
Layer   & 3.85/4 & 71.7 & 60 & 48min \\
Kernel  & 3.91/4 & 71.8 & 90 & 58min \\
\hline
\end{tabular}
\label{table:granularity}
\end{center}
\end{table}

As expected, while DBPs can be applied in different granularities, the layer-wise MPQ is still the optimal choice. When applying mixed precision to coarse-grained granularity, including the whole network and blocks, the performance of the trained MPQ network will be restricted because the difference of sensitivity within different components in the network is not fully utilized. Although searching the precision for finer-grained granularity such as different kernels will improve the accuracy of the MPQ model, the time cost for the strategy generation will grow significantly. Introducing more trainable parameters in the optimization process will result in more training instability. Moreover, current hardware accelerators rely highly on the parallelism of computing for the neural network. To the best of our knowledge, there are no kernel-wise hardware accelerators that successfully improve the efficiency in the real deployment of deep learning models. Based on these considerations, we focus more on layer-wise quantization when designing experiments.

\section{Training and Experimental Details} \label{sec:details}
\subsection{Training settings and hyper-parameters}
When comparing the results of our SDQ with other quantization methods in Table~\ref{tab:cifar} and Table~\ref{tab:imagenet}, we use the training settings and hyper-parameters shown in Table~\ref{tab:hyperpara}. Generally, most of these hyper-parameters are the same during the MPQ generation and MPQ training phase, while we observe that a larger batch size yields better performance during MPQ training.

\begin{table*}[h]
\centering
\caption{Detailed hyper-parameters and training scheme for different network architectures.}
\resizebox{\textwidth}{!}{
\begin{tabular}{c|cccccc}
\hline
Network             & \multicolumn{2}{c}{ResNet20 on CIFAR10}  & \multicolumn{2}{c}{ResNet18 on ImageNet-1K} & \multicolumn{2}{c}{MobileNetV2 on ImageNet-1K}    \\ 
\hline
Phase               & MPQ Generation    & MPQ Training  & MPQ Generation    & MPQ Training & MPQ Generation    & MPQ Training   \\
\hline
Epoch                & 100             &200               & 60          & 90             & 60                 &120          \\
Batch Size           & 512             &1024              & 256         & 512            & 256                &512          \\
Teacher              & -               & -                & ResNet101   & ResNet101      & ResNet101          &ResNet101    \\
Optimizer            & SGD             &SGD               & Adam        & Adam           & AdamW              &AdamW        \\
Initial \textit{lr}  & 0.1             &0.1               & 5e-4        & 1e-3           & 1e-3               & 1e-3  \\
\textit{lr} scheduler& MultiStepLR     &MultiStepLR       & Consine      & Consine       & Consine            & Consine      \\
Weight decay         & 1e-4            &1e-4              & -           & -              & -                  & -          \\
Warmup epochs        & -               &-                 & -            & -             & 30                 & 30   \\
\hline
Random Crop         & \checkmark      &\checkmark   & \checkmark      &\checkmark     & \checkmark      &\checkmark        \\
Random Flip         & \checkmark      &\checkmark   & \checkmark      &\checkmark     & \checkmark      &\checkmark      \\
Color jittering     & -               &-            & \checkmark      &\checkmark     & -               & -       \\
\hline
$\bm \lambda_Q$ in Eq.~\ref{eq:ebr}        & 1e-6         &-     & 1e-7        & -          & 1e-7        & -           \\
$\bm \lambda_E$ in Eq.~\ref{eq:qer}        & -            &5e-2  & -           & 1e-2       & -           & 1e-2       \\
$\bm \beta_{\text{thres}}$                 & 1e-4         &-     & 1e-5        & -          & 1e-5        & -          \\
Bitwidth candidate $\mathcal{B}$  & {1,2,3,4,5,6,7,8}&-   & {2,3,4,5,6,7,8} &- &{2,3,4,5,6,7,8}&- \\
\hline
\end{tabular}
}
\label{tab:hyperpara}
\end{table*}

\subsection{Training dynamics of SDQ models} 
\begin{figure}[t]
\centering
\subfigure{%
\includegraphics[width=0.4\textwidth]{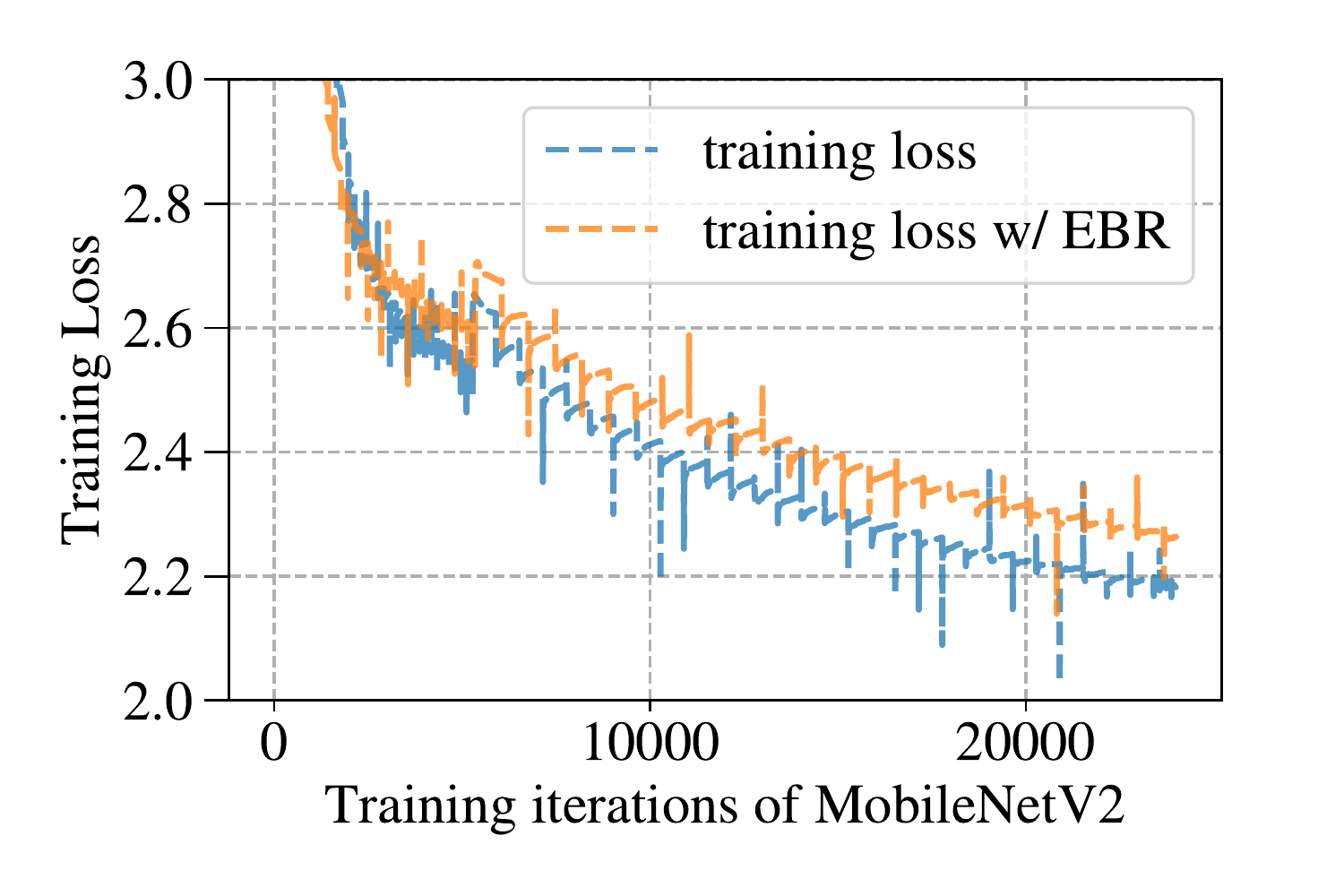}}%
\hspace{10mm}
\subfigure{%
\includegraphics[width=0.4\textwidth]{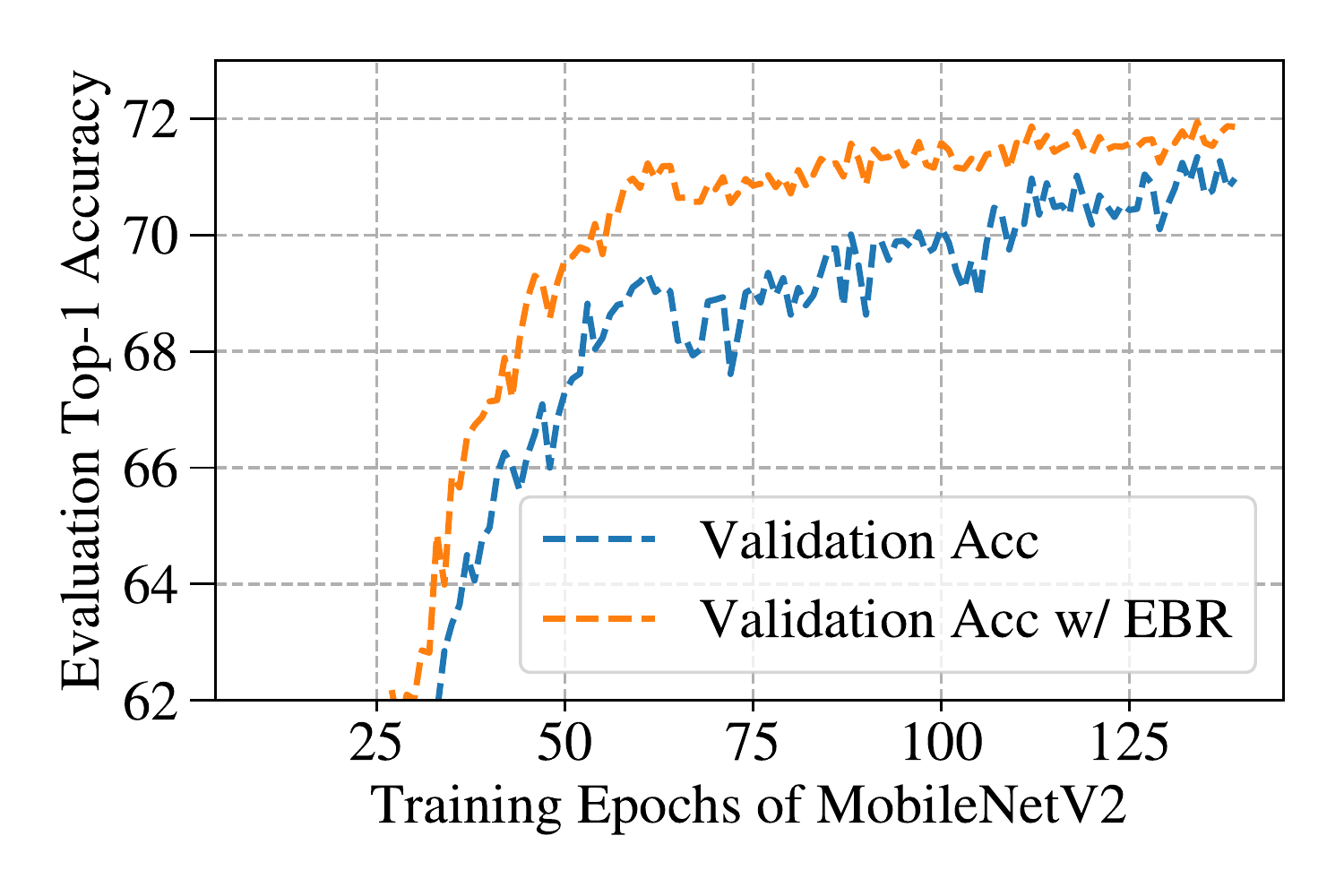}
}%
\caption{Training dynamics of SDQ with and without Entropy-aware Bin Regularization.}
\label{Figure:dynamics}
\end{figure}

Fig.~\ref{Figure:dynamics} illustrates the training loss and evaluation accuracy for our SDQ MobileNetV2 during the MPQ training phase. As shown in Fig.~\ref{fig:3dloss-linear}, quantized models are challenging to train as the latent non-smooth loss landscape makes models difficult to converge with inaccurately estimated gradients. We found that our proposed Entropy-aware Bin Regularization targeting to preserve more information can stabilize the training. Especially, when the learning rate drops during training, our EBR can effectively reduce the jitters in loss and accuracy values. 

\subsection{Detailed Comparison of Quantization Strategy}

\begin{figure}[H]
    \centering 
	\includegraphics[width=0.98\textwidth]{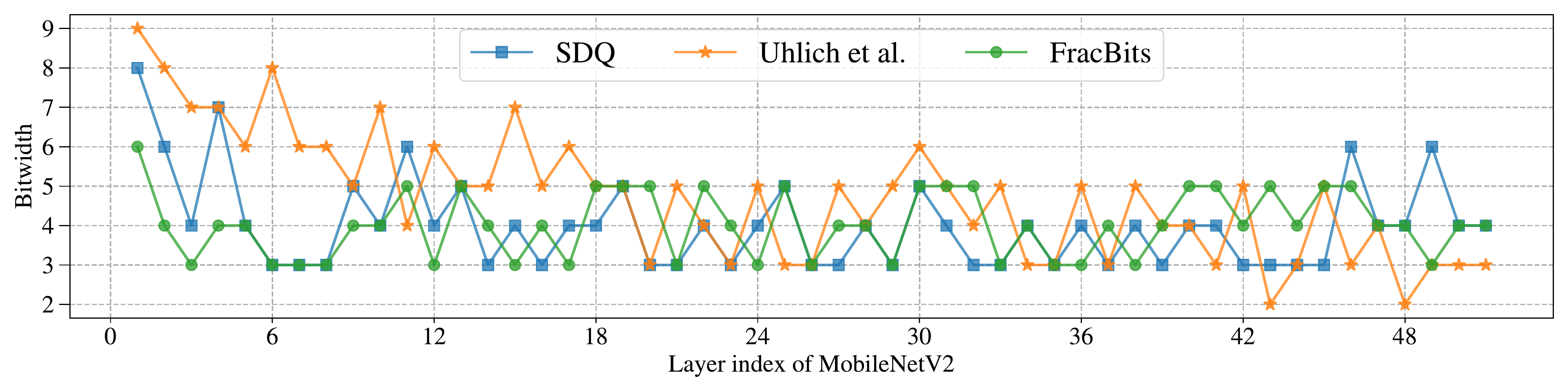}
	\vspace{-0.1in}
	\caption{Comparison of quantization strategy of all layers in MobileNetV2.}
	\label{Figure:bitw-compare}
\end{figure}

In Table~\ref{tab:strategy}, we compare our MobileNetV2 quantization strategy with Uhlich et al.~\cite{uhlich} and FracBits~\cite{yang2021fracbits} under the same initialization and training scheme. The results have demonstrated that our quantization is superior compared to previous strategies. All of the details on bitwidth assignment within each layer are shown in Fig.~\ref{Figure:bitw-compare}. We can see from the figure that all three bitwidth assignments follow similar patterns, such as the bitwidths for the first few layers are higher than average. Generally, layers with more parameters will have lower precision, while the strategy here yields different bitwidths for layers with the same number of parameters. This supports our statement in Sec.~\ref{sec:vis} that there is no simple heuristics to assign bitwidth based on some particular metrics directly. 

\end{document}